\documentclass[10pt,twocolumn,letterpaper]{article}

\usepackage{amsmath,amsfonts,bm}

\def\eqref#1{equation~\ref{#1}}

\def\1{\bm{1}}

\DeclareMathAlphabet{\mathsfit}{\encodingdefault}{\sfdefault}{m}{sl}
\SetMathAlphabet{\mathsfit}{bold}{\encodingdefault}{\sfdefault}{bx}{n}

\usepackage{cvpr}              
\usepackage{pifont}
\usepackage{booktabs}
\usepackage{wasysym}
\usepackage{graphicx}  
\usepackage{caption}   
\usepackage{lipsum}    
\usepackage{adjustbox}
\usepackage{multirow}

\usepackage[dvipsnames]{xcolor}

\definecolor{cvprblue}{rgb}{0.21,0.49,0.74}
\usepackage[pagebackref,breaklinks,colorlinks,citecolor=cvprblue]{hyperref}

\def\paperID{} 
\def\confName{CVPR}
\def\confYear{2026}

\title{Talking Together: Synthesizing Co-Located 3D Conversations from Audio}

\author{
Mengyi Shan$^{1}$, 
Shouchieh Chang$^{2}$, 
Ziqian Bai$^{2}$, 
Shichen Liu$^{2}$, 
Yinda Zhang$^{2}$, \\
Luchuan Song$^{3}$, 
Rohit Pandey$^{2}$, 
Sean Fanello$^{2}$, 
Zeng Huang$^{2}$ \\
\and
$^{1}$University of Washington, 
$^{2}$Google,
$^{3}$University of Rochester}

\begin{document}

\makeatletter
\let\@oldmaketitle\@maketitle
\renewcommand{\@maketitle}{\@oldmaketitle
}
\makeatother

\maketitle

\begin{abstract}
We tackle the challenging task of generating complete 3D facial animations for two interacting, co-located participants from a mixed audio stream. While existing methods often produce disembodied ``talking heads'' akin to a video conference call, our work is the first to explicitly model the dynamic 3D spatial relationship, including relative position, orientation, and mutual gaze, that is crucial for realistic in-person dialogues. Our system synthesizes the full performance of both individuals, including precise lip-sync, and uniquely allows their relative head poses to be controlled via textual descriptions. To achieve this, we propose a dual-stream architecture where each stream is responsible for one participant's output. We employ speaker's role embeddings and inter-speaker cross-attention mechanisms are designed to disentangle the mixed audio and model the interaction. Furthermore, we introduce a novel eye gaze loss to promote natural, mutual eye contact. To power our data-hungry approach, we introduce a novel pipeline to curate a large-scale conversational dataset consisting of over 2 million dyadic pairs from in-the-wild videos. Our method generates fluid, controllable, and spatially aware dyadic animations suitable for immersive applications in VR and telepresence, significantly outperforming existing baselines in perceived realism and interaction coherence.
\end{abstract}    
\vspace{-3mm}
\section{Introduction}
\label{sec:intro}

\begin{figure}
    \centering
    \includegraphics[width=\linewidth]{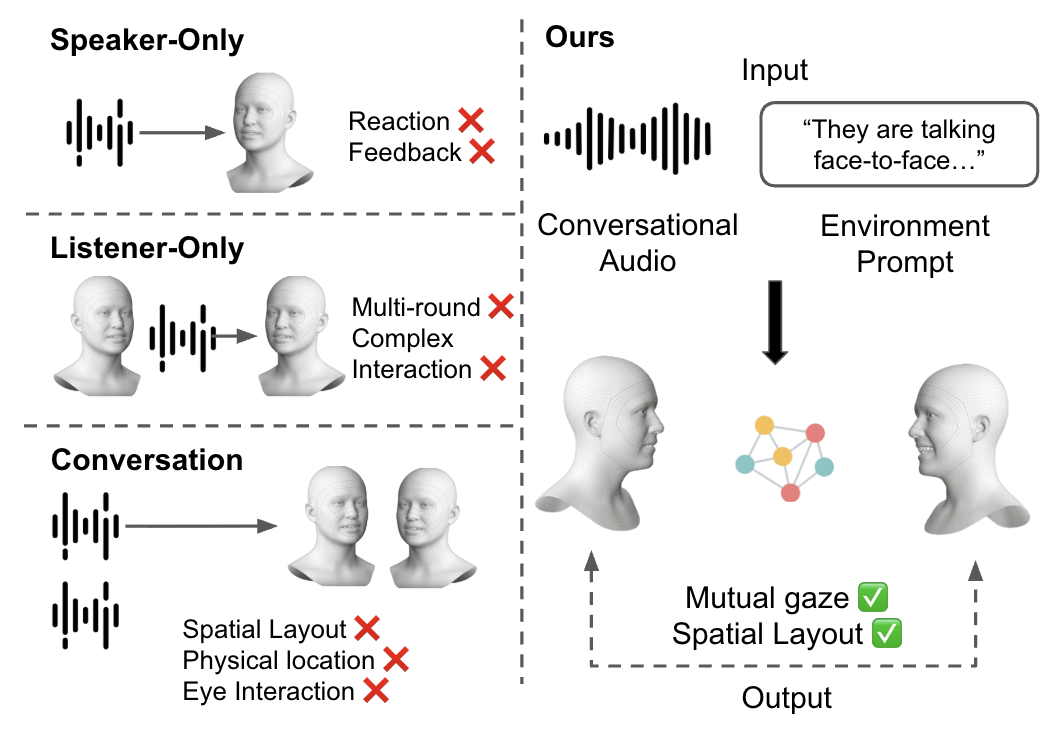}
    \caption{Our method in context.
Left: Prior work generates isolated participants, such as Speaker-Only models that lack listener reactions, Listener-Only models that do not model multi-round interactions, or Conversation models that resemble a ``video conference'' call, failing to model a shared 3D space, spatial layout, or real, physically meaningful eye interaction. 
Right: Our method takes a single mixed conversational audio stream and a text-based environment prompt to generate the complete, co-located 3D performance of \textit{both} participants. Our model is the first to explicitly synthesize this crucial spatial relationship, enabling realistic outputs with natural mutual gaze and a controllable spatial layout.
}
    \label{fig:teaser}
\vspace{-4mm}
\end{figure}

Photorealistic digital humans~\cite{cassell2000human,diederich2022design,liao2023proactive,wu2024vgg, sung2024laughtalk, li2023one, peng2024synctalk, niswar2009real,zhou2024meta} are foundational to immersive communication,  poised to revolutionize applications ranging from virtual reality and telepresence to advanced human-computer interfaces~\cite{zhen2023human,gowda2023pixels,yu2023nofa,pang2023dpe,bai2023high}. At the heart of believable human interaction is dyadic conversation, a complex and subtle dance of verbal and non-verbal signals exchanged between two individuals in a shared space. Current audio-driven 3D animation research either focuses on speaker-only generation~\cite{thambiraja2023imitator,xing2023codetalker, peng2023emotalk, peng2023selftalk, danecek2023emotional, kim2025memorytalker, thambiraja2025diface}, or overlooks the physical co-location of participants, predominantly generating isolated ``talking heads'' that resemble a video conference call rather than an in-person interaction~\cite{ng2023can,song2023react2023,liu2024customlistener,tran2024dyadic}. This paradigm fails to capture the crucial non-verbal cues derived from a shared 3D space, such as responsive head movements and mutual eye gaze that signal turn-taking and engagement. These methods ignore the spatial relationship between speakers and produce disconnected animations that lack the organic flow of a genuine face-to-face dialogue, presenting a barrier to truly immersive social interaction.

We present a method for generating complete 3D facial animations of two co-located participants, including their dynamic spatial relationship, from a single mixed audio stream as shown in Fig.~\ref{fig:teaser}. A significant barrier to progress in this domain is the scarcity of large-scale, high-fidelity 3D training data that captures natural dyadic conversations.

We address this challenge by effectively leveraging a combined corpus: a massive collection of over 2 million interacting pairs for learning general conversational dynamics, and a separate, high quality collection of a single-person talking head. The latter, sourced from segments with superior resolution and camera angles, is critical. It allows us to train a specialized lip-sync expert model, essential for remedying the inherent lip-sync ambiguities and artifacts often found in in-the-wild conversational data, thereby enabling our subsequent two-stage training strategy.


Leveraging these curated datasets, we introduce a novel generative model designed specifically for dyadic interaction. The method is centered on a dual-stream architecture, where each stream synthesizes the 3D facial performance for one participant. To effectively disentangle the single mixed audio input and model turn-taking dynamics, we employ cross-attention mechanisms and speaker role embeddings. This allows our model to learn not just an individual's speech patterns, but also the reactive, non-verbal behavior of the listener. This architecture is trained in a two-stage process that utilizes both of our datasets: we first pre-train the model's shared backbone on our diverse multi-person data to ensure expressive interactive behaviors, and then fine-tune the entire interactive model on the smaller scale high-quality mixed data to improve lip-sync.

Furthermore, we introduce two novel mechanisms to explicitly model the spatial dynamics of the conversation. First, to promote realistic mutual eye contact, we apply a targeted auxiliary eye gaze loss. We curate our dataset by filtering for a high-quality subset of interactions with clear and meaningful eye gaze, and this loss is applied exclusively when training on these selected samples. Second, we integrate a text-based control mechanism for the participants' relative positioning. During training, the model is conditioned on the ground-truth global translation of both heads in the first frame. At inference, this becomes a powerful control interface: a user can provide a natural language prompt (e.g., ``an intimate conversation'' or ``arguing across a table''), and a Large Language Model (LLM) predicts the initial 3D translations. This prediction is achieved in a few-shot manner, where the LLM is provided with several examples of text-to-translation pairs in its prompt.

Our key contributions can be summarized as follows:
\begin{itemize}
    \item We introduce an automated pipeline to curate a large-scale dataset of dyadic conversations from in-the-wild videos, along with a high-fidelity single-speaker corpus for robust lip-sync training.

    \item We propose a dual-stream diffusion architecture with a shared U-Net backbone, cross-attention, and FiLM conditioning to effectively model speaker interaction and disentangle a single mixed audio track.

    \item We employ a mixed-data training strategy, pre-training on massive real conversational data and then fine-tuning on high-quality, synthetic conversation data to ensure both precise lip articulation and natural, interactive behaviors.

    \item We enable intuitive scene control through a few-shot, LLM-based text-to-3D spatial translation mechanism and promote realistic eye contact with a targeted auxiliary gaze loss on a curated data subset.
\end{itemize}

\section{Related Works}
\label{sec:related_works}


\noindent \textbf{Generating Audio-based Talking Head. } 
Audio-driven talking head generation has emerged as a prominent area of research in recent years. One major line of work focuses on predicting 3D parametric face models~\cite{blanz1999morphable, flame2017}, which can be converted into mesh vertices for downstream applications such as virtual or augmented reality~\cite{aneja2023facetalk, sun2023vividtalk, nerffacespeech2024, yu2024gaussiantalker, lu2023facialanimation, meshtalk2021, faceformer2022, unitalker2024, facediffuser2023, emotalk2023, danecek2023emotional, kim2025memorytalker, thambiraja2025diface}. In parallel, another line directly synthesizes pixel-level outputs by animating a static portrait according to input audio, enabling more flexible and expressive media generation~\cite{min2022styletalker,xie2025julnet, su2024styleganedit, chopin2025dimitra, yi2020personalizedpose}. Recent advancements, such as FaceFormer~\cite{fan2022faceformer}, CodeTalker~\cite{xing2023codetalker}, and SelfTalk~\cite{peng2023selftalk}, introduced geometry-based methods using facial mesh representations to enhance realism in 3D talking heads. UniTalker~\cite{fan2024unitalker} improved generalization by training across multiple datasets and fine-tuning with minimal data, while ScanTalk~\cite{nocentini2024scantalk} enabled 3D face animation with any topology, thus broadening application scenarios. Our approach builds upon 3D-based generation to capture structural consistency, while further incorporating relative spatial information to enable co-located rendering.

\noindent \textbf{Generating Spatial-Aware Group Interactions. }
Recent research has progressed from generating isolated individuals to modeling multi-person interactions that respect spatial and social context. Early work explored multi-agent dynamics and collision-aware motion forecasting~\cite{tanke2023socialdiffusion, muller2024generativeproxemics}, while more recent methods generate coordinated and semantically grounded interactions driven by text or motion cues~\cite{shan2024opendomain, fan2024freemotion, tanaka2023roleaware, ghosh2023remos, sang2025timotion}. Several approaches further incorporate scene and contact awareness, synthesizing physically plausible movements conditioned on 3D environments or human–object interactions~\cite{cen2024text3dmotion, diller2023cghoi, ma2024contactaware}. Others model the communicative aspects of group behavior, generating expressive co-speech gestures and reactions within multi-person conversations~\cite{mughal2024convofusion}. Despite these advances, most existing frameworks focus on body dynamics and proxemics, with limited attention to integrating high-fidelity facial expressions and nuanced interpersonal cues, which is a key challenge for expressive, spatially aware group interaction generation.

\begin{figure*}[t]
    \centering
    \includegraphics[width=\linewidth]{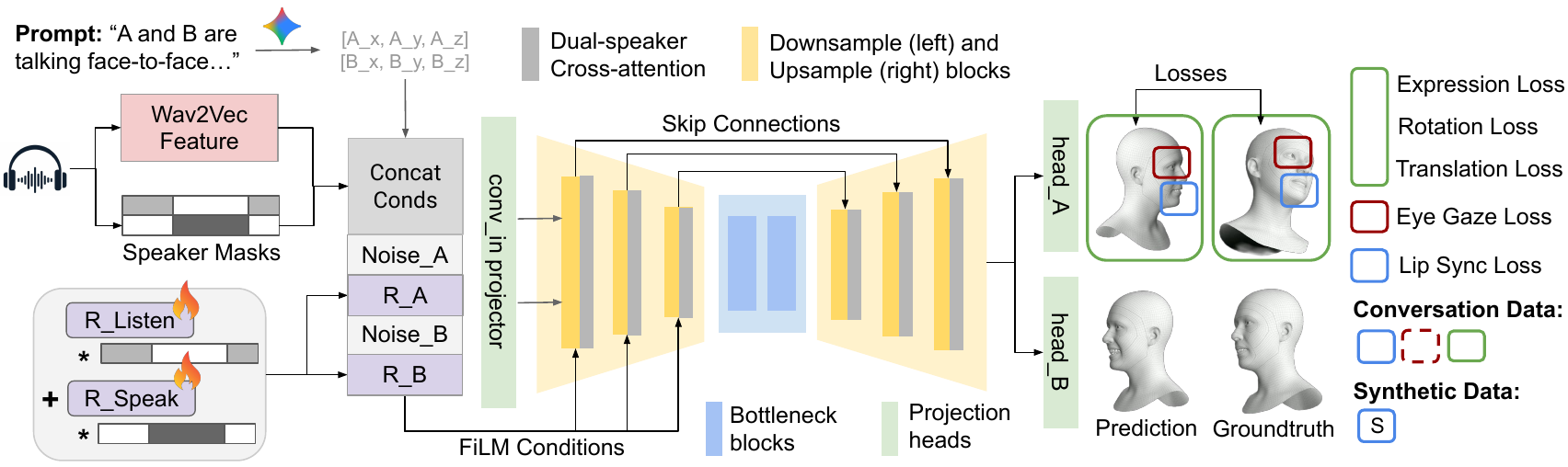}
    \caption{An overview of our dual-stream diffusion architecture. The model employs a shared U-Net backbone to process the noisy input streams for both participants in parallel. It is conditioned on features from the mixed audio, learnable role embeddings, and the speaker probability masks.  Dual-speaker cross-attention layers within the decoder allow the two streams to exchange information, modeling the interaction dynamics. The network outputs the predicted 3D animation parameters for expression ($\boldsymbol{\psi}$), rotation ($\boldsymbol{\theta}$)), and translation ($\boldsymbol{t}$). For conversation data, we utilize losses for all expression, rotation and translation prediction, with auxiliary eye gaze loss on samples with large-scale head movements. For synthetic data, we finetune with losses only on speaker's lip parameters. }
    \label{fig:pipeline}
\vspace{-3mm}
\end{figure*}

\noindent \textbf{Generating Conversation}
In human conversations, listeners convey vital non-verbal cues through facial expressions, head nods, and eye movements, which are crucial for natural and engaging interactions. Early works modeled such listener behaviors with neural networks~\cite{huang2019toward, zhou2022responsive, ng2023can, ng2024audio, luo2023reactface, liu2023mfr, song2023emotional, song2023react2023, liu2024customlistener, tran2024dyadic}. Zhou \etal~\cite{zhou2022responsive} introduced the task of responsive listener generation with a Role Switcher alternating between speaker and listener roles, while Learning2Listen~\cite{ng2022learning} produced brief reactions like nods and smiles from speech. In the video domain, INFP~\cite{zhu2024infp} learns conversational dynamics without explicit role assignments, and ARIG~\cite{guo2025arig} employs a long-range autoregressive model for flexible interactions. Closely related works include Audio2Photoreal~\cite{ng2024audio}, which generates photorealistic avatars from audio but lacks behavioral reactivity, and DualTalk~\cite{peng2025dualtalk}, which models dual-speaker conversations yet remains limited to ``video conferencing''-style talking heads without spatially aware co-located movements. In contrast, our system addresses these gaps by jointly learning implicit speaker–listener roles, for fine-grained mutual interaction and by being the first to synthesize all the participants within a shared, dynamic 3D spatial content.

\section{Method}

Our full pipeline is illustrated in Fig.~\ref{fig:pipeline}. We discuss our preliminaries, define the problem, and explain our model architecture as well as training paradigm below.

\subsection{Preliminary}

\label{sec:preliminary}

\noindent \textbf{Diffusion Model Objectives.}
We employ a denoising diffusion model that learns to reverse a fixed Gaussian noise process. Our model $G$, is trained to predict the original clean data $x_0$ from a noised input $x_t$ at timestep $t$, conditioned on a set of signals $c$. The objective is a simple L2 loss between the ground truth and the model's prediction:
\[\mathcal{L}=E_{x_0, t, x_t}\left[\left\|x_0-G\left(x_t, t, c\right)\right\|_2^2\right]\]
To enable classifier-free guidance during inference, we randomly mask the audio portion of the condition $c$ for 10\% of the training samples.

\noindent \textbf{Face Motion Representation.}
Our work utilizes a 3DMM-based parametric face model, where a 3D head mesh is defined by identity ($\boldsymbol{\beta}$), expression ($\boldsymbol{\psi}$), pose ($\boldsymbol{\theta}$), and translation ($\mathbf{t}$) parameters. The final facial geometry, $M(\boldsymbol{\beta}, \boldsymbol{\psi})$, is generated by adding a linear combination of identity and expression bases to a mean shape template ($\mathbf{S}$):
\begin{equation}
   M(\boldsymbol{\beta}, \boldsymbol{\psi}) = \mathbf{S} + \sum_{i} \beta_i \mathbf{S}_i + \sum_{j} \psi_j \mathbf{E}_j,
\end{equation}
where $\mathbf{S}_i$ and $\mathbf{E}_j$ are the principal components derived from a large dataset of high-quality 3D scans. This mesh is then articulated by pose and translation parameters using a standard Linear Blend Skinning rig to get animated geometry. 

\subsection{Problem Definition}
\label{sec:def}

Our primary objective is to synthesize a realistic 3D animation of a dyadic conversation from a single audio source. We formulate this as a sequence-to-sequence task where the input is a mixed audio waveform, $\mathbf{A} \in \mathbb{R}^{T}$, containing the speech of two participants. For each participant, the output is the concatenation of three sequences: a sequence of facial expression vectors, $\boldsymbol{\psi} \in \mathbb{R}^{L \times 63}$; a sequence of skeletal poses, $\boldsymbol{\theta} \in \mathbb{R}^{L \times 4 \times 3}$, representing the orientation of four key skeletal joints (neck, head, left and right eyes); and a sequence of head translations, $\mathbf{t} \in \mathbb{R}^{L \times 3}$, representing the global position in metric meters. The desired model output is a synchronized sequence of animation parameters of length $L$ for both participants, $\mathbf{x} = \{\mathbf{x_A}, \mathbf{x_B}\}$ where each $\mathbf{x} = \text{concat}(\mathbf{\psi}, \mathbf{\theta}, \mathbf{t}) \in \mathbb{R}^{L \times 78}$.

\subsection{Modeling Dyadic Interaction}
\label{sec:condition}
To model dyadic interaction, we propose a conditioned, dual-stream cross-attention diffusion architecture. 
We explain the modeling details below. 

\noindent \textbf{Audio Speaker Masking.} Given the mixed waveform, we predict two speaking probability masks 
$\mathbf{m}_A, \mathbf{m}_B \in [0,1]^{T\times1}$ 
corresponding to the speaking probability of each participant. 
These masks provide temporal guidance for disentangling speech and non-verbal behavior, and are used as conditioning inputs. Notably, the masks are not strictly accurate; this mild imperfection introduces beneficial noise that stabilizes training and improves robustness to real conversational overlap. At training time, we pre-computed and saved these masks instead of computing during the training loop (explained in Sec.~\ref{sec:data}). At inference time, we apply the audio separation and voice activity detection modules to estimates per-frame speaker activity on-the-fly.

\noindent \textbf{Shared Dual-Stream Architecture.}
We adopt a dual-stream design with shared weights across participants. 
A single U-Net backbone processes the noisy inputs $\mathbf{x_{t,A}}$ and $\mathbf{x_{t,B}}$ in parallel, promoting a unified, speaker-agnostic representation of facial motion. 
The shared generator $\mathcal{G}$ predicts denoised outputs for both streams as:
\begin{equation}
\hat{\mathbf{x}}_{0,A},\, \hat{\mathbf{x}}_{0,B} = 
\mathcal{G}\!\left(\mathbf{x}_{t,A},\, \mathbf{x}_{t,B},\, t,\, \mathbf{m}_A,\, \mathbf{m}_B\right),
\label{eq:dual_stream}
\end{equation}
This formulation ensures coherent evolution of both outputs while enabling downstream cross-attention layers to capture speaker–listener interactions.

\noindent \textbf{Interaction Modeling via Cross-Attention.}
To capture mutual influence between participants, we insert cross-attention layers within the shared U-Net decoder, enabling bidirectional information exchange. 
Let $\mathbf{h}_A$ and $\mathbf{h}_B$ denote the intermediate feature maps for the two streams. 
At each layer, the features of one participant are updated by attending to the other:
\begin{align}
\mathbf{h}'_A &= \mathrm{Attention}\!\left(\mathbf{Q}_A,\, \mathbf{K}_B,\, \mathbf{V}_B\right), \\
\mathbf{h}'_B &= \mathrm{Attention}\!\left(\mathbf{Q}_B,\, \mathbf{K}_A,\, \mathbf{V}_A\right),
\end{align}

where $\mathbf{Q}_A = \mathbf{W}_Q \mathbf{h}_A$, 
$\mathbf{K}_B = \mathbf{W}_K \mathbf{h}_B$, 
and $\mathbf{V}_B = \mathbf{W}_V \mathbf{h}_B$.
This symmetric exchange encourages each stream to model reactive, conversational behaviors while preserving speaker-specific representations.

\noindent \textbf{Speaker Role Embedding.}
To encode conversational intent, we introduce two learnable embedding vectors, $\mathbf{e}_\text{speak}$ and $\mathbf{e}_\text{listen}$, which are shared by both participants. At each animation timestep $k$, a dynamic role vector is calculated for each person by linearly interpolating these embeddings. The weighting $\mathbf{m}^{(k)}$ is the scalar speaker probability for that person at that frame, derived from their full speaker mask vector (e.g., $\mathbf{m}_A$).
\begin{equation}
    \mathbf{e}_{\text{role}}^{(k)} = \mathbf{m}^{(k)} \mathbf{e}_{\text{speak}} + (1 - \mathbf{m}^{(k)}) \mathbf{e}_{\text{listen}}.
\end{equation}
This resulting vector provides a continuous representation of a participant's interaction state. These role embeddings for both participants serve as key components of the model's overall conditioning signal, defined next.

\noindent \textbf{Dynamic Dual Conditioning.}
We condition the shared U-Net on a comprehensive set of multimodal cues capturing the conversational state and participant identity. These cues are consolidated into a single conditioning vector $\mathbf{c}^{(k)}$, which is dependent on the animation timestep $k$. This vector concatenates the Wav2Vec~\cite{baevski2020wav2vec} audio features $\mathbf{a}^{(k)}$, the dynamic role embeddings for both participants $\mathbf{e}_{\text{role}, A}^{(k)}$ and $\mathbf{e}_{\text{role}, B}^{(k)}$, and the speaker probability masks $\mathbf{m}_A^{(k)}$ and $\mathbf{m}_B^{(k)}$.
This vector $\mathbf{c}^{(k)}$ guides generation through two paths: (1) it is concatenated with the noisy latent input $\mathbf{x}$ before the diffusion U-Net, injecting global multimodal context, and (2) it modulates intermediate features $\mathbf{h}$ via FiLM:
\[
\mathrm{FiLM}(\mathbf{h}, \mathbf{c}^{(k)}) = (\boldsymbol{\gamma}(\mathbf{c}^{(k)}) + 1)\odot \mathbf{h} + \boldsymbol{\beta}(\mathbf{c}^{(k)}),
\]
where $\boldsymbol{\gamma}$ and $\boldsymbol{\beta}$ are learned functions that adaptively shift and rescale activations. These mechanisms jointly enable frame-dependent control over ``speaking'' and ``listening'' behaviors while preserving visual identity and coherence.

\subsection{Multi-Stage Training}

Conversational datasets exhibit poor lip synchronization due to low resolution and unavoidable occlusions as people might be facing each other in nature, and thus cannot be captured perfectly by the camera. To achieve accurate lip motion while retaining interactive dynamics, we adopt a two-stage training strategy with mixed datasets.

\noindent \textbf{Stage 1: Pre-training on Interaction Data.}  
We first pre-train the dual-stream model on large-scale conversational data to establish general audio–visual alignment and interaction understanding. This stage captures speaker–listener dynamics across diverse settings, learning rotation .

\noindent \textbf{Stage 2: Fine-tuning with High-Quality Lip Data.}  
Next, we fine-tune the model using a combination of (1) high-resolution single-speaker datasets with precise lip motion and (2) an augmented conversation data subset with high landmark confidence after super-resolution. For (1), we apply the L2 reconstruction loss only to the 20 parameters of the expression vector $\psi$ corresponding to lip and jaw articulation for the speaking participant, and set all other losses (e.g., rotation, translation, and non-lip expression) for both participants to zero. This staged process leverages large-scale conversational diversity for interaction grounding while preserving accurate lip synchronization through targeted fine-tuning on clean, high-quality visual data.

\subsection{Controllable Spatial Relationship}

To ground the conversation in 3D space and enable plausible interactions like eye contact, we explicitly condition the model on the participants' spatial layout.

\noindent \textbf{Training with Relative Translation.}
The model is trained to predict motion relative to a center-normalized starting position. At training time, we provide the ground-truth first-frame translations for both speakers, $\mathbf{t}_A^{(0)}$ and $\mathbf{t}_B^{(0)}$, as an additional conditioning signal. This static spatial layout is concatenated into the main conditioning vector $\mathbf{c}^{(k)}$, providing constant spatial context across all animation timesteps $k$. The model's prediction target for the translation component of $\mathbf{x}_0$ is then the \textit{delta} from this first frame: $\Delta\mathbf{t}^{(k)} = \mathbf{t}^{(k)} - \mathbf{t}^{(0)}$. This normalizes the problem, forcing the model to learn the movements including translation changes, head poses and eye rotations given absolute first frame position.

\noindent \textbf{Inference-Time Control via LLM.} This conditioning becomes a powerful control mechanism at inference. We leverage a Large Language Model (LLM)~\cite{reid2024gemini} to enable intuitive, text-based scene control. We employ a few-shot prompting strategy, where the LLM is provided with several human-annotated in-context examples of text-to-translation pairs (e.g., mapping a description like ``Standing side-by-side'' to a set of 3D coordinates for $t_A^{(0)}$ and $t_B^{(0)}$). This conditions the model to generate a structured coordinate set for a new, user-provided prompt. The full prompt structure and examples are detailed in the supplementary material.

\noindent \textbf{Auxiliary Eye Gaze Loss.}  
To promote realistic gaze behavior, our loss function computes the cosine similarity between the predicted gaze direction vector $\mathbf{g}_{pred}$ and the ground-truth gaze vector $\mathbf{g}_{gt}$. Specifically, we first convert the eye rotation parameters for each eye into a 3D gaze-forward vector. The final gaze direction vector, $g$, is then computed as the average of the left and right eye directions. This allows the model to learn the full spectrum of natural gaze, including both eye contact and aversion, by mimicking the ground truth. Crucially, we apply a higher loss weight \textit{selectively} to conversational samples that exhibit large-scale head movements (top 20\% head rotation variance of the dataset), which intuitively should exhibit more meaningful eye contact behaviors to learn from.

\subsection{Dataset Curation}
\label{sec:data}

\begin{figure}[t]
    \centering
    \includegraphics[width=\linewidth]{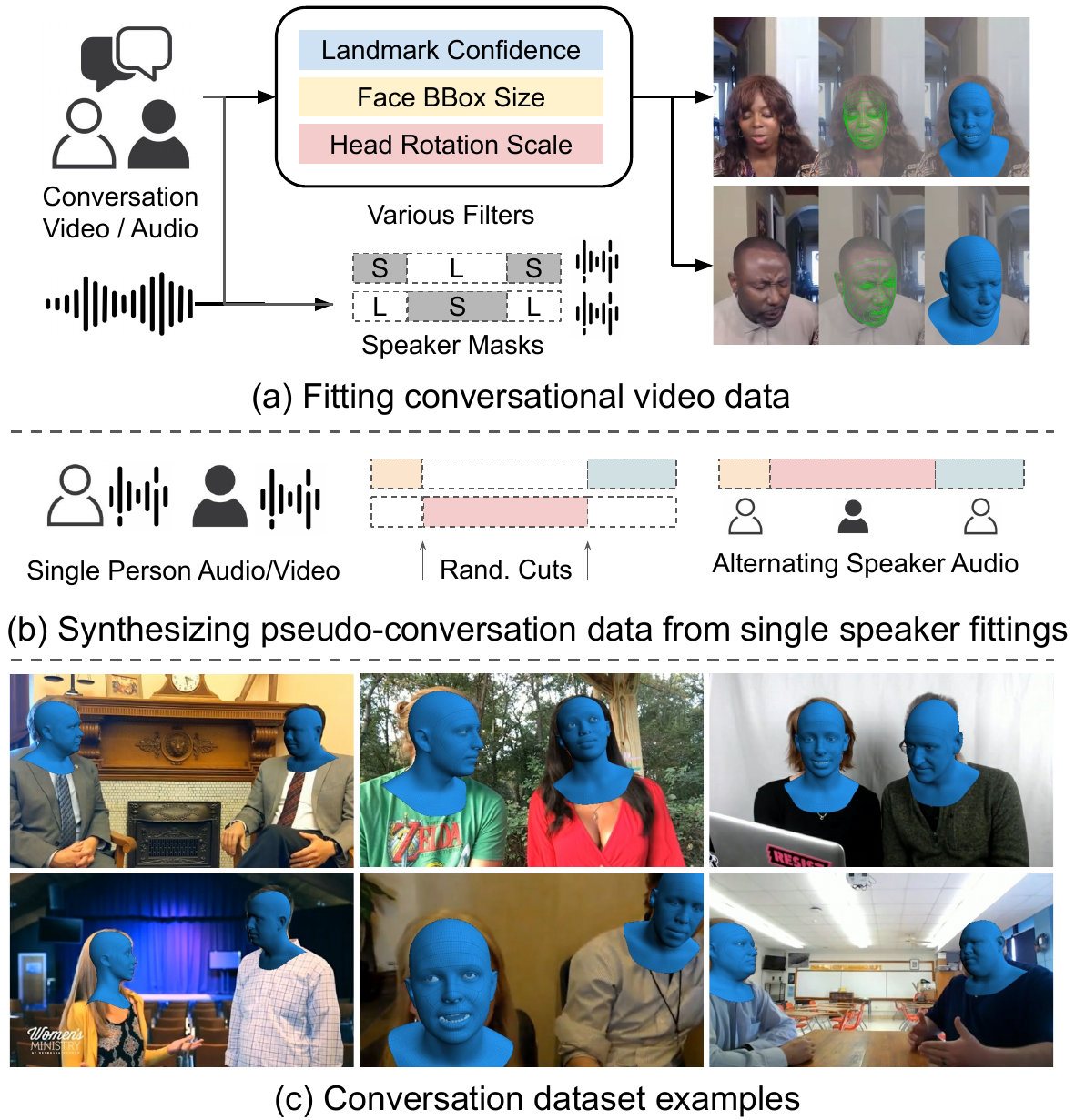}
    \caption{Our Data Curation Pipeline. This figure illustrates our two-pronged approach to dataset creation. 
    (a) Dyadic Conversation Dataset: We process raw conversational videos through various filters to reconstruct 3D facial parameters and extract speaker masks. 
    (b) Synthetic Dubbing Dataset: We generate clean pseudo-conversations by taking single-person videos, applying random cuts, and re-assembling them into new alternating-speaker audio tracks. 
    (c) Dataset Example: A sample of 3D reconstructions overlaid on videos from our curated dyadic dataset.}
    \label{fig:train-data}
\vspace{-3mm}
\end{figure}

\begin{figure*}[t]
    
    \begin{minipage}[t]{0.68\textwidth}
    \captionof{table}{%
            Comparison of 3D talking head datasets. Existing datasets differ in scale, interaction diversity, and lip accuracy. Our Dyadic Conversation and Synthetic Dubbing datasets combine large-scale interactive scenes with accurate lip motion and identity consistency, enabling joint learning of interaction and high-fidelity speech animation.
        }
        \begin{adjustbox}{width=\linewidth, valign=t}
            \begin{tabular}{@{}lcccccc@{}}
            \toprule
            \textbf{Datasets} & \textbf{Duration} & \textbf{Identities} & \textbf{Interaction} & \begin{tabular}[c]{@{}c@{}}\textbf{Multi-Round} \\ \textbf{Conversations}\end{tabular} & \begin{tabular}[c]{@{}c@{}}\textbf{Spatial} \\ \textbf{Relationship}\end{tabular} & \begin{tabular}[c]{@{}c@{}}\textbf{Simultaneous} \\ \textbf{Talking}\end{tabular} \\ \midrule
            VOCASET~\cite{cudeiro2019capture} & 0.5h & 12 & \ding{55} & \ding{55} & \ding{55} & \ding{55} \\
            BIWI~\cite{fanelli20103} & 1.44h & 14 & \ding{55} & \ding{55} & \ding{55} & \ding{55} \\
            ViCO~\cite{zhou2022responsive} & 1.6h & 92 & \ding{51} & \ding{55} & \ding{55} & \ding{55} \\
            L2L~\cite{ng2022learning} & 72h & 6 & \ding{51} & \ding{55} & \ding{55} & \ding{55} \\
            Lm\_listener~\cite{ng2023can} & 7h & 4 & \ding{51} & \ding{55} & \ding{55} & \ding{55} \\
            RealTalk~\cite{geng2023affective} & 8h & - & \ding{51} & \ding{55} & \ding{55} & \ding{55} \\
            DualTalk~\cite{peng2025dualtalk} & 50h & 1000+ & \ding{51} & \ding{51} & \ding{55} & \ding{55} \\ \midrule
            \textbf{Our Conversation} & \textbf{50,000+h} & \textbf{10k+} & \textbf{\ding{51}} & \textbf{\ding{51}} & \textbf{\ding{51}} & \ding{51} \\
            \textbf{Our Synthetic} & \textbf{50,000+h} & \textbf{10k+} & \textbf{\ding{55}} & \textbf{\ding{55}} & \textbf{\ding{55}} & \ding{51} \\ \bottomrule
            \end{tabular}
        \end{adjustbox}
        
        \label{tab:dataset_comparison}
    \end{minipage}%
    \hspace{0.02\textwidth}
    \begin{minipage}[t]{0.27\textwidth}
        \begin{adjustbox}{valign=t}
            \begin{minipage}{0.95\linewidth}
                \includegraphics[width=\linewidth]{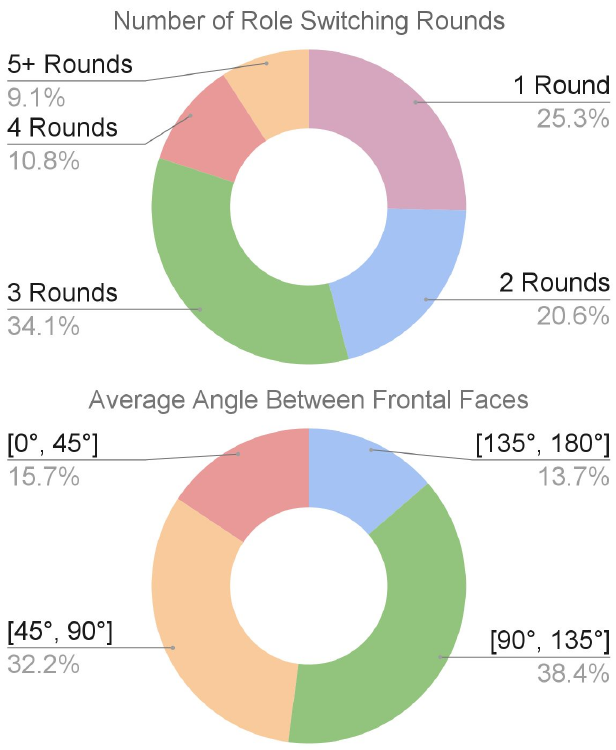}
                \captionof{figure}{%
                    Distribution visualization of the conversational dataset.
                }
                \label{fig:distribution}
            \end{minipage}
        \end{adjustbox}
    \end{minipage}
\vspace{-3mm}
\end{figure*}

A key challenge for training our model is the lack of a large-scale dataset containing 3D motion of co-located, interacting individuals. To address this, we curate two large-scale datasets with complementary strengths: a massive corpus of in-the-wild dyadic conversations for learning natural interactions, and a novel synthetic dubbing dataset with perfect ground truth for speaker activity and lip motion. Our full data generation pipeline is visualized in Fig.~\ref{fig:train-data}. We additionally compare our datasets with existing open-sourced datasets in Tab.~\ref{tab:dataset_comparison}, and visualize distribution characteristics of our conversational dataset in Fig.~\ref{fig:distribution}

\subsubsection{Dyadic Conversation Dataset}
This dataset captures complex non-verbal dynamics and spatial relationships of real-world interactions. Our data begins with a vast corpus of online conversational videos containing two people. Refer to Fig.~\ref{fig:train-data} (a) for the steps, and (c) for results overlaid on original videos.

\noindent \textbf{Scenario Filtering.}
We first automatically filter out scenarios where participants are not in the same physical space (e.g., video conference calls). We accomplished this by inspecting the background color distribution of left and right halves of the video, discarding videos that contain two separated scenes next to the middle seam.

\noindent \textbf{Quality Control Filtering.}
We aggressively filter the dataset by discarding videos where multiple frames contain heavily occluded, overly small or blurred faces. We take advantage of both the face bounding box sizes and the confidence values while predicting facial landmarks.

\noindent \textbf{Audio Source Separation and Masking.} 
We use Looking to Listen~\cite{ephrat2018l2l} as our source separation model to isolate each speaker's audio. We then incorporate voice activity detection based on WebRTC~\cite{ogura2021xvectorbasedvoiceactivity}, which generates binary speaker probability masks for each frame index.

\noindent \textbf{Video Quality Enhancement.}
We detect face bounding boxes and apply a face-specific super-resolution network similar to GLEAN~\cite{chan2022gleangenerativelatentbank} to specified face patches in the video. This enhances the clarity and detail of facial features. 

\noindent \textbf{3D Face Reconstruction.}
For each valid frame, we apply a robust 3D face reconstruction method to fit our parametric model to every detected face. This step yields per-frame expression parameters as well as the full 6DoF head pose, including global translation and orientation with respect to a fixed camera coordinate system for both faces. Absolute scales are estimated by assuming an average inter-ocular distance, allowing us to recover translation in approximate metric meters. We include a temporal term in the optimization process to ensure smooth transitioning between frames.

\subsubsection{Synthetic Dubbing Dataset}
While our dyadic dataset captures natural interactions, it presents two key limitations for high-fidelity training. First, it suffers from frequent occlusions; as participants naturally face each other, a single camera often fails to capture clear lip movements from all angles. Second, the speaker masks are imperfect, particularly in resolving ambiguous, overlapping speech. To generate a clean dataset with perfect ground-truth supervision, we introduce a synthetic dubbing pipeline. This pipeline uses a large corpus of high-quality, frontal-facing single-person videos, sourced from YouTube, which has highly accurate lip motion.

We randomly sample and cut speech segments from different videos, temporally aligning them to create synthetic two-person dialogues. By selectively muting one speaker while the other speaks, we simulate realistic conversational turn-taking. This ``dubbing'' process simulates complex scenarios, including simultaneous speech, with precision. Crucially, because the source for each speaker is isolated, we obtain perfect ground-truth speaker activity masks and inherit high-fidelity lip motion from the clean source data. This dataset is invaluable for training the model to handle overlapping audio and perform precise lip-sync. Fig.~\ref{fig:train-data} (b) shows the process of synthesizing pseudo-conversation data from single speaker fittings.

\section{Experiments}
\label{sec:experiments}

\begin{table*}[t]
\centering
\caption{Quantitative comparison of our method against baselines. We only compare translation results with retrieval-based methods, as other baselines inherently do not predict translations. Metrics: FD/P-FD (on renders) measure realism/interaction. MSE (on parameters) and vMSE (on vertices) measure accuracy for FULL motion, EXPression, ROTation, TRANSLation, EYE gaze, and LIP motion. SID measures diversity for LIStener and SPEaker roles. Lower is better ($\downarrow$) for FD, P-FD, MSE, and vMSE; higher is better ($\uparrow$) for SID. }
\resizebox{0.85\linewidth}{!}{
\begin{tabular}{@{}l|cc|ccccc|cc|cccc@{}}
\toprule
\multirow{2}{*}{\textbf{Methods}} 
& \multicolumn{1}{c|}{\textbf{FD} $\downarrow$} 
& \multicolumn{1}{c|}{\textbf{P-FD} $\downarrow$}
& \multicolumn{5}{c|}{\textbf{MSE} $\downarrow$} 
& \multicolumn{2}{c|}{\textbf{vMSE} $\downarrow$}
& \multicolumn{4}{c}{\textbf{SID} $\uparrow$} \\ 
\cmidrule(lr){2-3} \cmidrule(lr){4-8} \cmidrule(lr){9-10} \cmidrule(lr){11-14}
& \textbf{FULL} & \textbf{FULL}
& \textbf{EXP} & \textbf{TRANSL} & \textbf{ROT} & \textbf{EYE} & \textbf{LIP}
& \textbf{SPE} & \textbf{LIS}
& \textbf{FULL} & \textbf{EXP} & \textbf{ROT} & \textbf{TRANSL}\\
\midrule

CodeTalker~\cite{xing2023codetalker} & 47.23 & 70.54 & 10.47 & - & 14.28 & 3.07 & 2.95 & 12.49 & 6.85 & 0 & 0 & 0 & - \\
SelfTalk~\cite{peng2023selftalk}   & 43.58 & 53.98 & 8.21 & - & 11.59 & 2.47 & 2.41 & 10.98 & 6.13 & 1.68 & 1.27 & 1.39 & - \\
FaceFormer~\cite{faceformer2022} & 52.66 & 59.84 & 13.89 & - & 12.34 & 2.96 & 2.84 & 10.47 & 6.44 & 1.59 & 0.43 & 0.86 & - \\
Ours (Single) & 19.58 & 29.03 & 6.32 & - & 6.74 & 1.23 & 1.14 & \textbf{6.86} & 5.23 & 2.23 & 1.40 & 1.61 & - \\
\midrule
Listen-R   & 63.74 & 68.75 & 11.03 & 8.90 & 10.93 & 2.58 & 2.27 & 7.35 & 7.98 & 1.84 & 2.39 & 2.18 & \textbf{2.98} \\
Listen-A   & 65.09 & 41.57 & 12.57 & 7.67 & 7.98 & 2.32 & 1.91 & 7.68 & 4.89 & 1.29 & 2.26 & 2.57 & 1.96 \\
Listen-M   & 33.40 & 29.06 & 9.42 & 9.06 & 9.81 & 2.79 & 2.12 & 7.01 & 4.63 & 1.03 & 2.47 & \textbf{2.86} & 2.08 \\
\midrule
DIM~\cite{tran2023dim}        & 55.09 & 45.20 & 14.56 & - & 10.67 & 2.96 & 2.35 & 11.79 & 6.21 & 0.73 & 1.84 & 1.31 & - \\
DualTalk~\cite{peng2025dualtalk}   & 28.41 & 38.29 & 9.91 & - & 8.42 & 2.11 & 2.50 & 8.32 & 6.88 & 1.57 & 1.95  & 1.79 & - \\
L2L~\cite{ng2022learning} & 38.92 & 66.13 & 11.32 & - & 10.15 & 2.35 & 2.94 & 11.21 & 5.71 & 1.78 & 1.58 & 1.12 & - \\
\midrule
\textbf{Ours} & \textbf{10.43} & \textbf{18.24} 
& \textbf{4.03} & \textbf{2.09} & \textbf{3.50} & \textbf{0.98} & \textbf{0.35} 
& 7.99 & \textbf{2.29} & \textbf{2.28} & \textbf{2.48} & 1.97 & 2.45\\
\bottomrule
\end{tabular}}

\label{tab:comparison}
\end{table*}

\begin{figure*}
    \centering
    \includegraphics[width=0.95\linewidth]{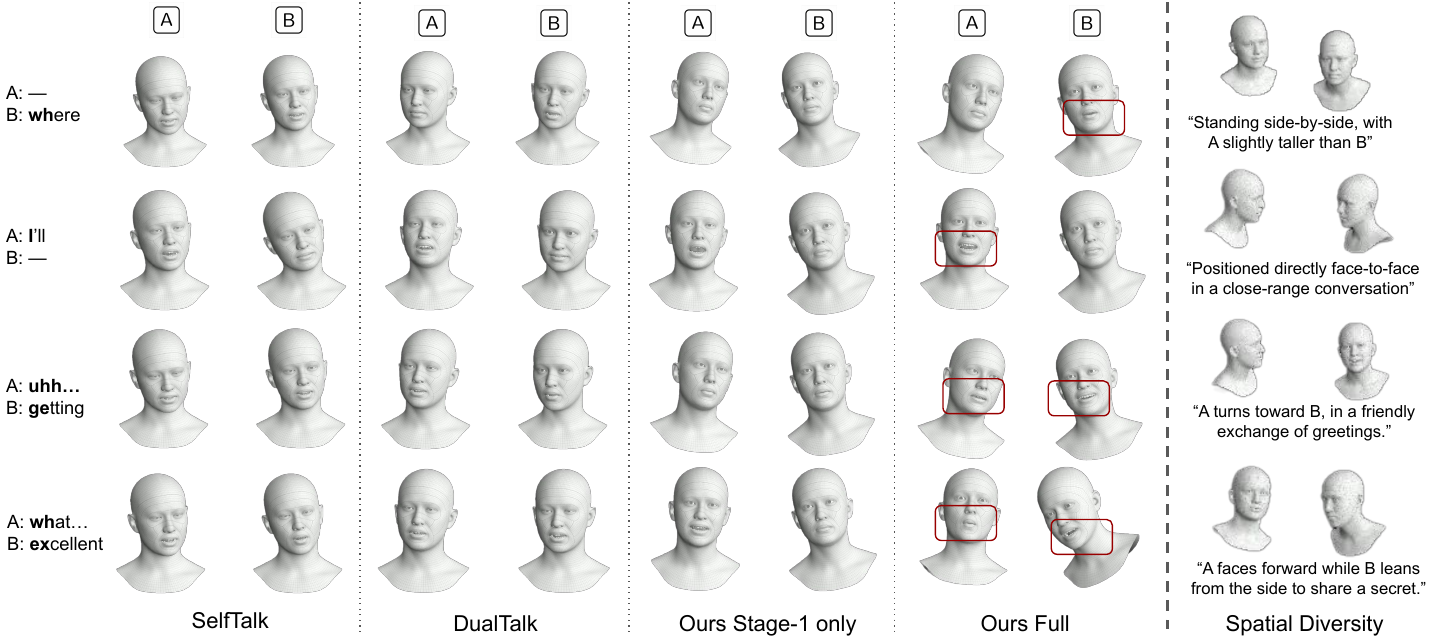}
    \caption{Left: Result of our model compared with baselines, shown as a few representative frames from one sequence. Top two rows feature speaker and listener interaction; Bottom two rows feature our model's ability to deal with short simultaneous speaking sequences. Right: Run inference with a variety of spatial layout prompts and render the first frame in a shared space. Red squares are used to mark significant speaker lip movements. This could be significantly easier to view as videos in supplementary. }
    \label{fig:result}
    \vspace{-4mm}
\end{figure*}

We present a thorough and detailed evaluation of our models below, including explanation of baselines and metrics. 

\subsection{Implementation details}
\label{sec:implementatin}

Our model is implemented in JAX. Audio is sampled at 16~kHz and converted to 768-dimensional Wav2Vec~2.0 features~\cite{baevski2020wav2vec2}. Training is performed on 16~A100 GPUs for 200{,}000 steps using AdamW, a cosine noise schedule, batch size $B=1024$, and learning rate $l=1\mathrm{e}{-4}$. At inference time we use a classifier-free guidance weight of 2.5. The total loss combines diffusion reconstruction losses on expression ($\lambda_{expr}\!=\!1$), rotation ($\lambda_{rot}\!=\!8$), and translation ($\lambda_{trans}\!=\!1$), with auxiliary regularization terms for vertex velocity ($\lambda_{vel}\!=\!1$) and eye gaze ($\lambda_{gaze}\!=\!5$). Audios are 10 seconds each, and all sequences are 25fps ($L=250$). Evaluation are performed on a test set of 2048 audio-face pairs not seen during training. Users can use Gemini~\cite{reid2024gemini} to predict initial 3D translations from text. For our quantitative results, we use the groundtruth translation as our condition to allow for per-vertex comparison with the test data. See Appendix for architecture and training details.

\subsection{Baselines}
\label{sec:baseline}

We compare our approach with three categories of baselines to evaluate different aspects of conversational modeling. 

\noindent \textbf{Single-Speaker Models.} We first compare with single-person talking head models that take clean audio as input. To apply them to conversational audio, we perform audio segmentation to obtain two separated audio tracks, which are then fed independently into each single-person model. These methods produce realistic speech-driven motion but lack any listener reaction or inter-speaker dependency. We include three such baselines with code available,  CodeTalker~\cite{xing2023codetalker}, FaceFormer~\cite{faceformer2022} and SelfTalk~\cite{peng2023selftalk}, and a single-person model trained on our single-person data. 

\noindent \textbf{Speaker + Listener Model.} We further evaluate baselines that combine a single-speaker motion generation model (for the active speaker) with separate listener motion models. We test our model on DIM~\cite{tran2023dim} and DualTalk~\cite{peng2025dualtalk}. We additionally compare with one random listener motion baseline (Listen-R) and two retrieval-based baselines that searches for a nearest-neighbor listener sequences from the training set based on either audio or speaker motion similarity (Listen-A, Listen-M). These comparisons assess how effectively each approach captures natural listener feedback without simultaneous modeling of interactions.  

\noindent \textbf{Two-Speaker Interaction Models.} Finally, we compare with L2L~\cite{ng2022learning}, which still doesn't generate both speakers simultaneously, but trains separate models for a specific identity that can generate either speaking or listening behaviors.


\subsection{Quantitative Evaluation}
\label{sec:quantitative}
We employ a similar set of metrics as DualTalk~\cite{peng2025dualtalk}. We omit the exact definition here and leave them to Appendix. 


\begin{table}[t]
\centering
\caption{Ablation study on different model components. 
Lower FD, P-FD, and MSE (EXPression, TRANslation or ROTation) indicate better performance when evaluated with groundtruth.}
\label{tab:ablation}
\resizebox{\linewidth}{!}{
\begin{tabular}{@{}l|c|c|ccc@{}} 
\toprule
\textbf{Ablation Setting} & \textbf{FD} $\downarrow$ & \textbf{P-FD} $\downarrow$ &
\textbf{EXP} $\downarrow$ & \textbf{TRAN} $\downarrow$ & \textbf{ROT} $\downarrow$ \\
\midrule
Single-Person Only & 50.45 & 50.05 & 10.01 & 2.32 & 3.88\\
w/o 2nd Stage & 60.12 & 64.44 & 7.73 & 1.71 & 2.94 \\
w/o Speaker Emb. & 35.92 & 35.93 & 7.18 & 1.80 & 2.87 \\
w/o Cross-Attn & 30.49 & 40.87 & 6.87 & 1.54 & 2.98\\
w/o Gaze Loss & 37.46 & 42.90 & 7.33 & 2.59 & 2.77\\
\midrule 
\textbf{Full Model (Ours)} & \textbf{21.71} & \textbf{22.56} & \textbf{5.97} & \textbf{1.50} & \textbf{2.48} \\
\bottomrule
\end{tabular}}
\vspace{-2mm}
\end{table}

\noindent \textbf{Baselines.} Quantitative comparison with baselines are shown in Table~\ref{tab:comparison}. Compared with all baselines, our method achieves the best performance across nearly all metrics on test sets, confirming the effectiveness of our system in capturing expressive, speaker-specific motion. Improvements in expression and pose MSE further demonstrate our cross-attention mechanism's performance in coordinating head motion between speakers. Our single person baseline slightly outperforms the conversation model on speaker vertex prediction, but worse in listener and interaction modeling. Retrieval-based methods occasionally generate more diverse rotations and translations but with low fidelity.

\noindent \textbf{Ablations.}  
We conduct ablations to assess each component's impact (Tab.~\ref{tab:ablation}). Training solely on single-person data performs poorly, as it provides no guidance for listener behavior. Conversely, training only on conversation data degrades facial expression accuracy, confirming this data source has high-quality interactions but low-quality expressions. Omitting the speaker-role embedding yields unnatural listener responses. Removing cross-attention fragments turn-taking and increases FD/P-FD, confirming its role in modeling interaction. Finally, excluding the eye-gaze loss reduces mutual attention realism and increases both P-FD and expression MSE. These results demonstrate that all of our datasets, training stages, and architecture jointly enhance quantitative accuracy and perceptual quality.

\subsection{Qualitative Evaluation}
\label{sec:qualitative}
We prepare extensive qualitative results showing our model's performance comparing with baselines (single speaker and speaker + listener) and alternative designs (without the second stage). We refer the readers to the videos in the supplementary files for illustration. In Fig.~\ref{fig:result}, we present generated frames with a single input audio . 

\noindent \textbf{Conversation Generation.}
Top two rows in Fig~\ref{fig:result} visualize complete conversational sequences generated from mixed audio. Our method produces smooth, coherent facial motion and clear turn-taking behavior. Bottom two rows in Fig~\ref{fig:result} show that our model disentangles overlapping speech and generates plausible mouth motion for both speakers. When both participants speak concurrently, each stream maintains accurate lip articulation and independent head motion while preserving the conversational context. 

\noindent \textbf{LLM-guided Spatial Relationship Control.}
Right column in Fig.~\ref{fig:result} demonstrates the controllability of relative positioning through text prompts. Our system composes 3D layouts while maintaining realistic head translation and gaze behavior. Our model also consistently produces mutual eye contact and context-aware gaze shifts. While text-to-translation is not the focus of our evaluation, we provide a full user study on its effectiveness in the Appendix.

\subsection{Human Evaluation}
\label{sec:human_eval}

\begin{table}[t]
  \centering
  \caption{Human evaluation preference rate (\%). Higher is better. Participants selected the best model from four options.}
  \label{tab:human_eval_brief}
  \resizebox{0.96\linewidth}{!}{
  \begin{tabular}{@{}l|ccccc@{}} 
    \toprule
    \textbf{Method} & \textbf{Lip} & \textbf{Speaker M} & \textbf{Listener M} & \textbf{Interact} & \textbf{Gaze} \\
    \midrule
    SelfTalk    & 0.9 & 0.9 & 1.6 & 1.6 & 2.4 \\
    DualTalk    & 3.9 & 6.3 & 7.2 & 5.6 & 7.9 \\
    Ours (S1)      & 15.9 & 19.0 & 18.2 & 21.4 & 21.4 \\
    \textbf{Ours} & \textbf{79.3} & \textbf{73.8} & \textbf{73.0} & \textbf{71.4} & \textbf{68.3} \\
    \bottomrule
  \end{tabular}
  }
  \vspace{-3mm}
\end{table}
Automatic metrics cannot fully capture the naturalness and mutual responsiveness of co-located conversational animation, so we performed a forced-choice user study. Nineteen Participants were shown 14 groups of clips of two-person conversations. Specifically, we test it on our model, our model with first stage only, the best conversation model DualTalk~\cite{peng2025dualtalk} and the best single person model SelfTalk~\cite{peng2023selftalk}. For each clip, participants were asked to select the single best result according to five aspects:

\begin{itemize}
    \item \textbf{Speaker Lip Quality:} The realism and accuracy of the lip-sync for whoever is actively speaking.
    \item \textbf{Speaker Movements:} The naturalness and vividness of the speaker's head movements (excluding the lip region).
    \item \textbf{Listener Movements:} The realism and appropriateness of the listener's head movements and expression.
    \item \textbf{Interaction Quality:} The naturalness of the non-verbal interaction between the two subjects, including mutual head turning, reactions, etc.
    \item \textbf{Eye Gaze Quality:} The realism of eye movements for both participants during communication.
\end{itemize}

As shown in Tab.~\ref{tab:human_eval_brief}, our model was selected as the best most frequently across all criteria. It outperformed baselines significantly in terms of interaction and eye gaze. Training with the second stage provides greatly improves lip sync, also aligning with the quantitative metrics.

\section{Conclusion}

We address the challenging task of generating complete 3D facial animations for two co-located, interacting participants from a single mixed audio stream. Our framework explicitly models the dynamic 3D spatial relationship between speakers, a crucial yet often overlooked aspect of in-person dialogue. Enabled by a large-scale dataset built through in-the-wild video curation and a synthetic dubbing pipeline, our dual-stream generative model with cross-attention and an eye gaze loss produces realistic, controllable, and spatially coherent dyadic interactions.


{
    \small
    \bibliographystyle{ieeenat_fullname}
    \bibliography{main}

@inproceedings{cudeiro2019capture,
  title={Capture, learning, and synthesis of 3D speaking styles},
  author={Cudeiro, Daniel and Bolkart, Timo and Laidlaw, Cassidy and Ranjan, Anurag and Black, Michael J},
  booktitle={Proceedings of the IEEE/CVF conference on computer vision and pattern recognition},
  pages={10101--10111},
  year={2019}
}

@inproceedings{fan2022faceformer,
  title={Faceformer: Speech-driven 3d facial animation with transformers},
  author={Fan, Yingruo and Lin, Zhaojiang and Saito, Jun and Wang, Wenping and Komura, Taku},
  booktitle={Proceedings of the IEEE/CVF Conference on Computer Vision and Pattern Recognition},
  pages={18770--18780},
  year={2022}
}

@inproceedings{thambiraja2023imitator,
  title={Imitator: Personalized speech-driven 3d facial animation},
  author={Thambiraja, Balamurugan and Habibie, Ikhsanul and Aliakbarian, Sadegh and Cosker, Darren and Theobalt, Christian and Thies, Justus},
  booktitle={Proceedings of the IEEE/CVF International Conference on Computer Vision},
  pages={20621--20631},
  year={2023}
}

@inproceedings{xing2023codetalker,
  title={Codetalker: Speech-driven 3d facial animation with discrete motion prior},
  author={Xing, Jinbo and Xia, Menghan and Zhang, Yuechen and Cun, Xiaodong and Wang, Jue and Wong, Tien-Tsin},
  booktitle={Proceedings of the IEEE/CVF Conference on Computer Vision and Pattern Recognition},
  pages={12780--12790},
  year={2023}
}

@inproceedings{peng2023emotalk,
  title={Emotalk: Speech-driven emotional disentanglement for 3d face animation},
  author={Peng, Ziqiao and Wu, Haoyu and Song, Zhenbo and Xu, Hao and Zhu, Xiangyu and He, Jun and Liu, Hongyan and Fan, Zhaoxin},
  booktitle={Proceedings of the IEEE/CVF International Conference on Computer Vision},
  pages={20687--20697},
  year={2023}
}

@inproceedings{peng2023selftalk,
  title={Selftalk: A self-supervised commutative training diagram to comprehend 3d talking faces},
  author={Peng, Ziqiao and Luo, Yihao and Shi, Yue and Xu, Hao and Zhu, Xiangyu and Liu, Hongyan and He, Jun and Fan, Zhaoxin},
  booktitle={Proceedings of the 31st ACM International Conference on Multimedia},
  pages={5292--5301},
  year={2023}
}

@inproceedings{ng2022learning,
  title={Learning to listen: Modeling non-deterministic dyadic facial motion},
  author={Ng, Evonne and Joo, Hanbyul and Hu, Liwen and Li, Hao and Darrell, Trevor and Kanazawa, Angjoo and Ginosar, Shiry},
  booktitle={Proceedings of the IEEE/CVF Conference on Computer Vision and Pattern Recognition},
  pages={20395--20405},
  year={2022}
}

@inproceedings{song2023emotional,
  title={Emotional listener portrait: Neural listener head generation with emotion},
  author={Song, Luchuan and Yin, Guojun and Jin, Zhenchao and Dong, Xiaoyi and Xu, Chenliang},
  booktitle={Proceedings of the IEEE/CVF International Conference on Computer Vision},
  pages={20839--20849},
  year={2023}
}

@inproceedings{ng2023can,
  title={Can language models learn to listen?},
  author={Ng, Evonne and Subramanian, Sanjay and Klein, Dan and Kanazawa, Angjoo and Darrell, Trevor and Ginosar, Shiry},
  booktitle={Proceedings of the IEEE/CVF International Conference on Computer Vision},
  pages={10083--10093},
  year={2023}
}

@inproceedings{song2023react2023,
  title={React2023: The first multiple appropriate facial reaction generation challenge},
  author={Song, Siyang and Spitale, Micol and Luo, Cheng and Barquero, Germ{\'a}n and Palmero, Cristina and Escalera, Sergio and Valstar, Michel and Baur, Tobias and Ringeval, Fabien and Andr{\'e}, Elisabeth and others},
  booktitle={Proceedings of the 31st ACM International Conference on Multimedia},
  pages={9620--9624},
  year={2023}
}

@inproceedings{liu2024customlistener,
  title={CustomListener: Text-guided Responsive Interaction for User-friendly Listening Head Generation},
  author={Liu, Xi and Guo, Ying and Zhen, Cheng and Li, Tong and Ao, Yingying and Yan, Pengfei},
  booktitle={Proceedings of the IEEE/CVF Conference on Computer Vision and Pattern Recognition},
  pages={2415--2424},
  year={2024}
}

@inproceedings{ng2024audio,
  title={From audio to photoreal embodiment: Synthesizing humans in conversations},
  author={Ng, Evonne and Romero, Javier and Bagautdinov, Timur and Bai, Shaojie and Darrell, Trevor and Kanazawa, Angjoo and Richard, Alexander},
  booktitle={Proceedings of the IEEE/CVF Conference on Computer Vision and Pattern Recognition},
  pages={1001--1010},
  year={2024}
}

@inproceedings{pang2023dpe,
  title={Dpe: Disentanglement of pose and expression for general video portrait editing},
  author={Pang, Youxin and Zhang, Yong and Quan, Weize and Fan, Yanbo and Cun, Xiaodong and Shan, Ying and Yan, Dong-ming},
  booktitle={Proceedings of the IEEE/CVF Conference on Computer Vision and Pattern Recognition},
  pages={427--436},
  year={2023}
}

@inproceedings{bai2023high,
  title={High-fidelity facial avatar reconstruction from monocular video with generative priors},
  author={Bai, Yunpeng and Fan, Yanbo and Wang, Xuan and Zhang, Yong and Sun, Jingxiang and Yuan, Chun and Shan, Ying},
  booktitle={Proceedings of the IEEE/CVF Conference on Computer Vision and Pattern Recognition},
  pages={4541--4551},
  year={2023}
}

@inproceedings{yu2023nofa,
  title={Nofa: Nerf-based one-shot facial avatar reconstruction},
  author={Yu, Wangbo and Fan, Yanbo and Zhang, Yong and Wang, Xuan and Yin, Fei and Bai, Yunpeng and Cao, Yan-Pei and Shan, Ying and Wu, Yang and Sun, Zhongqian and others},
  booktitle={ACM SIGGRAPH 2023 conference proceedings},
  pages={1--12},
  year={2023}
}

@inproceedings{huang2019toward,
  title={Toward rnn based micro non-verbal behavior generation for virtual listener agents},
  author={Huang, Hung-Hsuan and Fukuda, Masato and Nishida, Toyoaki},
  booktitle={Social Computing and Social Media. Design, Human Behavior and Analytics: 11th International Conference, SCSM 2019, Held as Part of the 21st HCI International Conference, HCII 2019, Orlando, FL, USA, July 26-31, 2019, Proceedings, Part I 21},
  pages={53--63},
  year={2019},
  organization={Springer}
}

@inproceedings{zhou2022responsive,
  title={Responsive listening head generation: a benchmark dataset and baseline},
  author={Zhou, Mohan and Bai, Yalong and Zhang, Wei and Yao, Ting and Zhao, Tiejun and Mei, Tao},
  booktitle={European Conference on Computer Vision},
  pages={124--142},
  year={2022},
  organization={Springer}
}

@article{luo2023reactface,
  title={Reactface: Multiple appropriate facial reaction generation in dyadic interactions},
  author={Luo, Cheng and Song, Siyang and Xie, Weicheng and Spitale, Micol and Shen, Linlin and Gunes, Hatice},
  journal={arXiv preprint arXiv:2305.15748},
  year={2023}
}

@inproceedings{liu2023mfr,
  title={Mfr-net: Multi-faceted responsive listening head generation via denoising diffusion model},
  author={Liu, Jin and Wang, Xi and Fu, Xiaomeng and Chai, Yesheng and Yu, Cai and Dai, Jiao and Han, Jizhong},
  booktitle={Proceedings of the 31st ACM International Conference on Multimedia},
  pages={6734--6743},
  year={2023}
}

@article{tran2024dyadic,
  title={Dyadic Interaction Modeling for Social Behavior Generation},
  author={Tran, Minh and Chang, Di and Siniukov, Maksim and Soleymani, Mohammad},
  journal={arXiv preprint arXiv:2403.09069},
  year={2024}
}

@article{baevski2020wav2vec,
  title={wav2vec 2.0: A framework for self-supervised learning of speech representations},
  author={Baevski, Alexei and Zhou, Yuhao and Mohamed, Abdelrahman and Auli, Michael},
  journal={Advances in neural information processing systems},
  volume={33},
  pages={12449--12460},
  year={2020}
}

@inproceedings{sung2024laughtalk,
  title={Laughtalk: Expressive 3d talking head generation with laughter},
  author={Sung-Bin, Kim and Hyun, Lee and Hong, Da Hye and Nam, Suekyeong and Ju, Janghoon and Oh, Tae-Hyun},
  booktitle={Proceedings of the IEEE/CVF Winter Conference on Applications of Computer Vision},
  pages={6404--6413},
  year={2024}
}

@inproceedings{niswar2009real,
  title={Real-time 3D talking head from a synthetic viseme dataset},
  author={Niswar, Arthur and Ong, Ee Ping and Nguyen, Hong Thai and Huang, Zhiyong},
  booktitle={Proceedings of the 8th International Conference on Virtual Reality Continuum and its Applications in Industry},
  pages={29--33},
  year={2009}
}

@inproceedings{li2023one,
  title={One-shot high-fidelity talking-head synthesis with deformable neural radiance field},
  author={Li, Weichuang and Zhang, Longhao and Wang, Dong and Zhao, Bin and Wang, Zhigang and Chen, Mulin and Zhang, Bang and Wang, Zhongjian and Bo, Liefeng and Li, Xuelong},
  booktitle={Proceedings of the IEEE/CVF Conference on Computer Vision and Pattern Recognition},
  pages={17969--17978},
  year={2023}
}

@inproceedings{peng2024synctalk,
  title={Synctalk: The devil is in the synchronization for talking head synthesis},
  author={Peng, Ziqiao and Hu, Wentao and Shi, Yue and Zhu, Xiangyu and Zhang, Xiaomei and Zhao, Hao and He, Jun and Liu, Hongyan and Fan, Zhaoxin},
  booktitle={Proceedings of the IEEE/CVF Conference on Computer Vision and Pattern Recognition},
  pages={666--676},
  year={2024}
}

@article{cassell2000human,
  title={Human conversation as a system framework: Designing embodied conversational agents},
  author={Cassell, Justine and Bickmore, Tim and Campbell, Lee and Vilhjalmsson, Hannes and Yan, Hao and others},
  journal={Embodied conversational agents},
  pages={29--63},
  year={2000}
}

@article{diederich2022design,
  title={On the design of and interaction with conversational agents: An organizing and assessing review of human-computer interaction research},
  author={Diederich, Stephan and Brendel, Alfred Benedikt and Morana, Stefan and Kolbe, Lutz},
  journal={Journal of the Association for Information Systems},
  volume={23},
  number={1},
  pages={96--138},
  year={2022}
}

@inproceedings{liao2023proactive,
  title={Proactive conversational agents in the post-chatgpt world},
  author={Liao, Lizi and Yang, Grace Hui and Shah, Chirag},
  booktitle={Proceedings of the 46th International ACM SIGIR Conference on Research and Development in Information Retrieval},
  pages={3452--3455},
  year={2023}
}

@article{zhen2023human,
  title={Human-computer interaction system: A survey of talking-head generation},
  author={Zhen, Rui and Song, Wenchao and He, Qiang and Cao, Juan and Shi, Lei and Luo, Jia},
  journal={Electronics},
  volume={12},
  number={1},
  pages={218},
  year={2023},
  publisher={MDPI}
}

@article{gowda2023pixels,
  title={From pixels to portraits: A comprehensive survey of talking head generation techniques and applications},
  author={Gowda, Shreyank N and Pandey, Dheeraj and Gowda, Shashank Narayana},
  journal={arXiv preprint arXiv:2308.16041},
  year={2023}
}

@article{fan2024unitalker,
  title={UniTalker: Scaling up Audio-Driven 3D Facial Animation through A Unified Model},
  author={Fan, Xiangyu and Li, Jiaqi and Lin, Zhiqian and Xiao, Weiye and Yang, Lei},
  journal={arXiv preprint arXiv:2408.00762},
  year={2024}
}

@article{nocentini2024scantalk,
  title={ScanTalk: 3D Talking Heads from Unregistered Scans},
  author={Nocentini, Federico and Besnier, Thomas and Ferrari, Claudio and Arguillere, Sylvain and Berretti, Stefano and Daoudi, Mohamed},
  journal={arXiv preprint arXiv:2403.10942},
  year={2024}
}

@article{geng2023affective,
  title={Affective faces for goal-driven dyadic communication},
  author={Geng, Scott and Teotia, Revant and Tendulkar, Purva and Menon, Sachit and Vondrick, Carl},
  journal={arXiv preprint arXiv:2301.10939},
  year={2023}
}

@article{fanelli20103,
  title={A 3-d audio-visual corpus of affective communication},
  author={Fanelli, Gabriele and Gall, Juergen and Romsdorfer, Harald and Weise, Thibaut and Van Gool, Luc},
  journal={IEEE Transactions on Multimedia},
  volume={12},
  number={6},
  pages={591--598},
  year={2010},
  publisher={IEEE}
}

@article{wu2024vgg,
  title={VGG-Tex: A Vivid Geometry-Guided Facial Texture Estimation Model for High Fidelity Monocular 3D Face Reconstruction},
  author={Wu, Haoyu and Peng, Ziqiao and Zhou, Xukun and Cheng, Yunfei and He, Jun and Liu, Hongyan and Fan, Zhaoxin},
  journal={arXiv preprint arXiv:2409.09740},
  year={2024}
}

@article{zhou2024meta,
  title={Meta-Learning Empowered Meta-Face: Personalized Speaking Style Adaptation for Audio-Driven 3D Talking Face Animation},
  author={Zhou, Xukun and Li, Fengxin and Peng, Ziqiao and Wu, Kejian and He, Jun and Qin, Biao and Fan, Zhaoxin and Liu, Hongyan},
  journal={arXiv preprint arXiv:2408.09357},
  year={2024}
}

@article{aneja2023facetalk,
  title={FaceTalk: Audio-driven motion diffusion for neural parametric head models},
  author={Aneja, Shivangi and Thies, Justus and Dai, Angela and Nie{\ss}ner, Matthias},
  journal={arXiv preprint arXiv:2312.08459},
  year={2023}
}

@article{sun2023vividtalk,
  title={VividTalk: One-shot audio-driven talking head generation based on {3d} hybrid prior},
  author={Sun, Xusen and Zhang, Longhao and Zhu, Hao and Zhang, Peng and Zhang, Bang and Ji, Xinya and Zhou, Kangneng and Gao, Daiheng and Bo, Liefeng and Cao, Xun},
  journal={arXiv preprint arXiv:2312.01841},
  year={2023}
}

@article{nerffacespeech2024,
  title={{NeRFFaceSpeech}: One-shot audio-driven {3d} talking head synthesis via generative priors + {nerf}},
  author={Anonymous Author(s)},
  journal={arXiv preprint arXiv:2405.05749},
  year={2024}
}

@article{yu2024gaussiantalker,
  title={GaussianTalker: Speaker-specific talking head synthesis via {3d} gaussian splatting},
  author={Yu, Hongyun and Qu, Zhan and Yu, Qihang and Chen, Jianchuan and Jiang, Zhonghua and Chen, Zhiwen and Zhang, Shengyu and Xu, Jimin and Wu, Fei and Lv, Chengfei and Yu, Gang},
  journal={arXiv preprint arXiv:2404.14037},
  year={2024}
}

@article{lu2023facialanimation,
  title={Audio-driven {3d} facial animation from in-the-wild videos},
  author={Lu, Liying and Zhang, Tianke and Liu, Yunfei and Chu, Xuangeng and Li, Yu},
  journal={arXiv preprint arXiv:2306.11541},
  year={2023}
}

@article{min2022styletalker,
  title={StyleTalker: One-shot style-based audio-driven talking head video generation},
  author={Min, Dongchan and Song, Minyoung and Ko, Eunji and Hwang, Sung Ju},
  journal={arXiv preprint arXiv:2208.10922},
  year={2022}
}

@article{xie2025julnet,
  title={Audio-driven talking face video generation with joint uncertainty learning},
  author={Xie, Yifan and Ma, Fei and Bin, Yi and He, Ying and Yu, Fei},
  journal={arXiv preprint arXiv:2504.18810},
  year={2025}
}

@article{su2024styleganedit,
  title={Audio-driven high-resolution seamless talking head video editing via {stylegan}},
  author={Su, Jiacheng and Liu, Kunhong and Chen, Liyan and Yao, Junfeng and Liu, Qingsong and Lv, Dongdong},
  journal={arXiv preprint arXiv:2407.05577},
  year={2024}
}

@article{chopin2025dimitra,
  title={Dimitra: Audio-driven diffusion model for expressive talking head generation},
  author={Chopin, Baptiste and Dhamija, Tashvik and Balaji, Pranav and Wang, Yaohui and Dantcheva, Antitza},
  journal={arXiv preprint arXiv:2502.17198},
  year={2025}
}

@article{yi2020personalizedpose,
  title={Audio-driven talking face video generation with personalized head pose},
  author={Yi, Ran and Ye, Zipeng and Zhang, Juyong and Bao, Hujun and Liu, Yong-Jin},
  journal={arXiv preprint arXiv:2002.10137},
  year={2020}
}

@article{flame2017,
  title={FLAME: Learning a model of facial shape and expression from 4D scans},
  author={Lüthi, Marcel and Gerig, Thomas and Jud, Clément and Vetter, Thomas},
  journal={ACM Transactions on Graphics (TOG)},
  volume={36},
  number={6},
  pages={1--17},
  year={2017},
  publisher={ACM},
  doi={10.1145/3130800.3130813},
  archivePrefix={arXiv},
  eprint={1709.08506}
}

@inproceedings{meshtalk2021,
  title={MeshTalk: 3{D} face animation from speech using cross-modality disentanglement},
  author={Richard, Alexander and Zollhöfer, Michael and Wen, Yandong and de la Torre, Fernando and Sheikh, Yaser},
  booktitle={Proceedings of the IEEE/CVF International Conference on Computer Vision},
  pages={5782--5791},
  year={2021}
}

@inproceedings{faceformer2022,
  title={FaceFormer: Speech-driven 3{D} facial animation with transformers},
  author={Fan, Yingruo and Lin, Zhaojiang and Saito, Jun and Wang, Wenping and Komura, Taku},
  booktitle={Proceedings of the IEEE/CVF Conference on Computer Vision and Pattern Recognition},
  pages={18770--18780},
  year={2022}
}

@inproceedings{unitalker2024,
  title={UniTalker: Scaling up audio-driven 3{D} facial animation through a unified model},
  author={Fan, Xiangyu and Li, Jiaqi and Lin, Zhiqian and Xiao, Weiye and Yang, Lei},
  booktitle={Proceedings of the European Conference on Computer Vision},
  year={2024}
}

@inproceedings{emotalk2023,
  title={EmoTalk: Speech-driven emotional disentanglement for 3{D} face animation},
  author={Peng, Ziqiao and others},
  booktitle={Proceedings of the IEEE/CVF International Conference on Computer Vision},
  year={2023}
}

@article{facediffuser2023,
  title={FaceDiffuser: Speech-driven 3{D} facial animation synthesis using diffusion},
  author={Stan, Stefan and Haque, Kazi Injamamul and Yumak, Zerrin},
  journal={Proceedings of ACM on Multimedia Interactive Experiences (MIG)},
  year={2023}
}

@inproceedings{blanz1999morphable,
  title={A morphable model for the synthesis of 3d faces},
  author={Blanz, Volker and Vetter, Thomas},
  booktitle={Proceedings of the 26th annual conference on Computer graphics and interactive techniques},
  pages={187--194},
  year={1999}
}

@inproceedings{peng2025dualtalk,
    title={DualTalk: Dual-Speaker Interaction for 3D Talking Head Conversations},
    author={Ziqiao Peng and Yanbo Fan and Haoyu Wu and Xuan Wang and Hongyan Liu and Jun He and Zhaoxin Fan},
    booktitle={Proceedings of the IEEE/CVF Conference on Computer Vision and Pattern Recognition},
    year={2025},
}

@article{zhu2024infp,
      title={INFP: Audio-driven interactive head generation in dyadic conversations},
      author={Zhu, Yongming and Zhang, Longhao and Rong, Zhengkun and Hu, Tianshu and Liang, Shuang and Ge, Zhipeng},
      journal={arXiv preprint arXiv:2412.04037},
      year={2024}
}

@misc{guo2025arig,
      title={ARIG: Autoregressive Interactive Head Generation for Real-time Conversations}, 
      author={Ying Guo and Xi Liu and Cheng Zhen and Pengfei Yan and Xiaoming Wei},
      year={2025},
      eprint={2507.00472},
      archivePrefix={arXiv},
      primaryClass={cs.CV},
      url={https://arxiv.org/abs/2507.00472}, 
}

@inproceedings{tran2023dim,
author = {Tran, Minh and Chang, Di and Siniukov, Maksim and Soleymani, Mohammad},
title = {DIM: Dyadic Interaction Modeling for Social Behavior Generation},
year = {2024},
isbn = {978-3-031-72912-6},
publisher = {Springer-Verlag},
address = {Berlin, Heidelberg},
url = {https://doi.org/10.1007/978-3-031-72913-3_27},
doi = {10.1007/978-3-031-72913-3_27},
booktitle = {Computer Vision – ECCV 2024: 18th European Conference, Milan, Italy, September 29–October 4, 2024, Proceedings, Part XXXVII},
pages = {484–503},
numpages = {20},
location = {Milan, Italy}
}

@inproceedings{shan2024opendomain,
  title     = {Towards Open-Domain Text-Driven Synthesis of Multi-Person Motions},
  author    = {Mengyi Shan and Lu Dong and Yutao Han and Yuan Yao and Tao Liu and Ifeoma Nwogu and Guo-Jun Qi and Mitch Hill},
  booktitle = {European Conference on Computer Vision (ECCV)},
  year      = {2024},
  url       = {https://arxiv.org/abs/2405.18483}
}

@inproceedings{fan2024freemotion,
  title     = {FreeMotion: A Unified Framework for Number-Free Text-to-Motion Synthesis},
  author    = {Ke Fan and Junshu Tang and Weijian Cao and Ran Yi and Moran Li and Jingyu Gong and Jiangning Zhang and Yabiao Wang and Chengjie Wang and Lizhuang Ma},
  booktitle = {European Conference on Computer Vision (ECCV)},
  year      = {2024},
  url       = {https://arxiv.org/abs/2405.15763}
}

@inproceedings{ghosh2023remos,
  title     = {ReMoS: 3D Motion-Conditioned Reaction Synthesis for Two-Person Interactions},
  author    = {Anindita Ghosh and Rishabh Dabral and Vladislav Golyanik and Christian Theobalt and Philipp Slusallek},
  booktitle = {European Conference on Computer Vision (ECCV)},
  year      = {2024},
  url       = {https://arxiv.org/abs/2311.17057}
}

@inproceedings{sang2025timotion,
  title     = {TIMotion: Temporal and Interactive Modeling for Human-Human Motion Generation},
  author    = {Shihao Sang and Xiaokang Yang and Yujin Chen and Xiaoyi Dong and Yujun Cai and Yong-Jin Liu},
  booktitle = {Conference on Computer Vision and Pattern Recognition (CVPR)},
  year      = {2025},
  url       = {https://arxiv.org/abs/2408.17135}
}

@inproceedings{tanaka2023roleaware,
  title     = {Role-Aware Interaction Generation from Textual Description},
  author    = {Mikihiro Tanaka and Kent Fujiwara},
  booktitle = {International Conference on Computer Vision (ICCV)},
  year      = {2023},
  url       = {https://openaccess.thecvf.com/content/ICCV2023/papers/Tanaka_Role-Aware_Interaction_Generation_from_Textual_Description_ICCV_2023_paper.pdf}
}

@inproceedings{tanke2023socialdiffusion,
  title     = {Social Diffusion: Long-term Multiple Human Motion Anticipation},
  author    = {Jan Tanke and Yi Zhou and Mohamed Omran and others},
  booktitle = {International Conference on Computer Vision (ICCV)},
  year      = {2023},
  url       = {https://openaccess.thecvf.com/content/ICCV2023/papers/Tanke_Social_Diffusion_Long-term_Multiple_Human_Motion_Anticipation_ICCV_2023_paper.pdf}
}

@inproceedings{muller2024generativeproxemics,
  title     = {Generative Proxemics: A Prior for 3D Social Interaction from Images},
  author    = {Lea Müller and Vickie Ye and Georgios Pavlakos and Michael J. Black and Angjoo Kanazawa},
  booktitle = {Conference on Computer Vision and Pattern Recognition (CVPR)},
  year      = {2024},
  url       = {https://arxiv.org/abs/2306.09337}
}

@inproceedings{mughal2024convofusion,
  title     = {ConvoFusion: Multi-Modal Conversational Diffusion for Co-Speech Gesture Synthesis},
  author    = {Muhammad Hamza Mughal and Rishabh Dabral and Ikhsanul Habibie and Lucia Donatelli and Marc Habermann and Christian Theobalt},
  booktitle = {Conference on Computer Vision and Pattern Recognition (CVPR)},
  year      = {2024},
  url       = {https://arxiv.org/abs/2403.17936}
}

@inproceedings{cen2024text3dmotion,
  title     = {Generating Human Motion in 3D Scenes from Text Descriptions},
  author    = {Zhi Cen and Huaijin Pi and Sida Peng and Zehong Shen and Minghui Yang and Shuai Zhu and Hujun Bao and Xiaowei Zhou},
  booktitle = {Conference on Computer Vision and Pattern Recognition (CVPR)},
  year      = {2024},
  url       = {https://openaccess.thecvf.com/content/CVPR2024/papers/Cen_Generating_Human_Motion_in_3D_Scenes_from_Text_Descriptions_CVPR_2024_paper.pdf}
}

@inproceedings{diller2023cghoi,
  title     = {CG-HOI: Contact-Guided 3D Human-Object Interaction Generation},
  author    = {Christian Diller and Angela Dai},
  booktitle = {Conference on Computer Vision and Pattern Recognition (CVPR)},
  year      = {2023},
  url       = {https://arxiv.org/abs/2311.16097}
}

@inproceedings{ma2024contactaware,
  title     = {Contact-aware Human Motion Generation from Textual Descriptions},
  author    = {Sihan Ma and Qiong Cao and Jing Zhang and Dacheng Tao},
  booktitle = {arXiv preprint},
  year      = {2024},
  url       = {https://arxiv.org/abs/2403.15709}
}

@inproceedings{baevski2020wav2vec2,
  title     = {wav2vec 2.0: A Framework for Self-Supervised Learning of Speech Representations},
  author    = {Baevski, Alexei and Zhou, Henry and Mohamed, Abdelrahman and Auli, Michael},
  booktitle = {Advances in Neural Information Processing Systems (NeurIPS)},
  year      = {2020},
  pages     = {12449--12460}
}

@article{reid2024gemini,
  title   = {Gemini 1.5: Unlocking multimodal understanding across modalities},
  author  = {Reid, M. and Anil, R. and Badia, A. and et~al.},
  journal = {arXiv preprint arXiv:2403.05530},
  year    = {2024}
}

@article{ephrat2018l2l,
   title={Looking to listen at the cocktail party: a speaker-independent audio-visual model for speech separation},
   volume={37},
   ISSN={1557-7368},
   url={http://dx.doi.org/10.1145/3197517.3201357},
   DOI={10.1145/3197517.3201357},
   number={4},
   journal={ACM Transactions on Graphics},
   publisher={Association for Computing Machinery (ACM)},
   author={Ephrat, Ariel and Mosseri, Inbar and Lang, Oran and Dekel, Tali and Wilson, Kevin and Hassidim, Avinatan and Freeman, William T. and Rubinstein, Michael},
   year={2018},
   month=jul, pages={1–11} }

@misc{ogura2021xvectorbasedvoiceactivity,
      title={X-Vector based voice activity detection for multi-genre broadcast speech-to-text}, 
      author={Misa Ogura and Matt Haynes},
      year={2021},
      eprint={2112.05016},
      archivePrefix={arXiv},
      primaryClass={eess.AS},
      url={https://arxiv.org/abs/2112.05016}, 
}

@misc{chan2022gleangenerativelatentbank,
      title={GLEAN: Generative Latent Bank for Image Super-Resolution and Beyond}, 
      author={Kelvin C. K. Chan and Xiangyu Xu and Xintao Wang and Jinwei Gu and Chen Change Loy},
      year={2022},
      eprint={2207.14812},
      archivePrefix={arXiv},
      primaryClass={cs.CV},
      url={https://arxiv.org/abs/2207.14812}, 
}

@inproceedings{danecek2023emotional, series={SA ’23},
   title={Emotional Speech-Driven Animation with Content-Emotion Disentanglement},
   url={http://dx.doi.org/10.1145/3610548.3618183},
   DOI={10.1145/3610548.3618183},
   booktitle={SIGGRAPH Asia 2023 Conference Papers},
   publisher={ACM},
   author={Daněček, Radek and Chhatre, Kiran and Tripathi, Shashank and Wen, Yandong and Black, Michael and Bolkart, Timo},
   year={2023},
   month=dec, pages={1–13},
   collection={SA ’23} }

@article{kim2025memorytalker,
        author = {Kim, Hyung Kyu and Lee, Sangmin and Kim, Hak Gu}, 
        title = {MemoryTalker: Personalized Speech-Driven 3D Facial Animation via Audio-Guided Stylization},
        journal = {Proceedings of the IEEE/CVF International Conference on Computer Vision (ICCV)}, 
        month = {October},
        year = {2025}
        }

@inproceedings{
      thambiraja2025diface,
      title={3Di{FACE}: Synthesizing and Editing Holistic 3D Facial Animation},
      author={Balamurugan Thambiraja and Malte Prinzler and Sadegh Aliakbarian and Darren Cosker and Justus Thies},
      booktitle={International Conference on 3D Vision 2025},
      year={2025},
      url={https://openreview.net/forum?id=8qpjYG1x8I}
      }
}

\clearpage
\setcounter{page}{1}
\maketitlesupplementary
\setcounter{section}{0}

\section{Video Results}
To demonstrate the effectiveness of our framework, we provide a folder containing 14 distinct conversational sequences sourced from our test set. This can be best viewed by clicking on the \texttt{video\_results.html} file with your favorite browser (Chrome recommended!) These examples cover diverse dynamics, including rapid turn-taking and overlapping speech. For each sequence, we compare four methods:

\begin{itemize}
    \item \textbf{Ours (Full):} Our complete dual-stream model.
    \item \textbf{Ours (Stage 1):} Ablation without the second-stage high-fidelity lip fine-tuning.
    \item \textbf{DualTalk:} The state-of-the-art dual-speaker baseline.
    \item \textbf{SelfTalk:} Single-speaker baseline applied to each track.
\end{itemize}

We encourage observation from the following aspects:
\begin{itemize}
    \item \textbf{Listener Responsiveness:} Ours generates natural reactions (nods, eye movements) while listening, whereas SelfTalk often remains frozen.
    \item \textbf{Interaction Coherence:} Ours maintains spatial awareness and mutual gaze, avoiding the disconnected ``video conference'' feel of DualTalk.
    \item \textbf{Lip-Sync Precision:} Comparing Ours (Full) vs. Ours (Stage 1) highlights the necessity of our second-stage fine-tuning for sharp, accurate articulation.
\end{itemize}

\section{Implementation Details}
\subsection{Latent Diffusion Model}
Our model is an audio-driven facial animation system based on a latent diffusion model, designed to generate sequences of parameters for a 3D parametric face model, including expression, joint rotations, and root translation. The generation is conditioned on pre-trained audio embeddings and facial identity. The core network is a U-Net with an embedding dimension of 512. This U-Net consists of 2 main blocks, each with 2 subblocks. The second block incorporates self-attention with 8 heads and a temporal stride of 2. The model predicts parameters for a reduced PCA version of the face model, encompassing 63 dimensions for expression, 50 for identity, 12 for joint rotations (representing 4 joints as axis-angle vectors), and 3 for root translation.  The model processes sequences of 250 frames. We employ classifier-free guidance with a conditioning signal dropout probability of 0.1 and a guidance scale of 2.0 during inference, which utilizes 8 DDIM sampling steps. The diffusion process is trained to predict the clean data ($x_0$) using a cosine-based logSNR schedule. The model is trained using the Adafactor optimizer with a learning rate of 1e-4, which remains constant after a 2000-step linear warmup, and a batch size of 1024. The loss function is a weighted sum of L2 losses on the face model parameters with specific setup and weights described in the paper. 

\subsection{Few-Shot Spatial Layout Generation}
\label{sec:layout}

To enable text-driven control over the initial spatial arrangement of the participants, we utilize \textbf{Gemini 2.5 Pro} via API. We employ a few-shot prompting strategy to condition the model to output precise 3D translation vectors $t \in \mathbb{R}^3$ (in metric meters) for both Speaker A and Speaker B.

\noindent\textbf{Prompt Structure.} The prompt consists of three parts: (1) a system instruction defining the coordinate system (where $+Z$ is forward and $+X$ is right) and the origin point (center of the interaction); (2) three in-context examples demonstrating diverse spatial relationships; and (3) the user's input query. We enforce a strict JSON output format to ensuring programmatic parsability.

\noindent\textbf{Inference Parameters.} We set the temperature to $0.2$ to encourage deterministic and stable coordinate predictions. The maximum output token limit is set to 128.

\noindent\textbf{Prompt Template.} Below is an example structure of the prompt provided to the LLM. In practice, we manually label 30 examples, and randomly choose 3 as few-shot examples in each inference round. 

\begin{quote}
\small
\texttt{System: You are a 3D scene layout assistant. Generate 3D head translation coordinates (in meters) for two people (A and B) based on a text description. The center of the conversation is (0,0,0). Output ONLY valid JSON.}

\vspace{0.5em}
\texttt{[Example 1]} \\
\texttt{Input: "Standing face-to-face in a normal conversation."} \\
\texttt{Output: \{"A": [0.0, 0.0, -0.5], "B": [0.0, 0.0, 0.5]\}}

\vspace{0.5em}
\texttt{[Example 2]} \\
\texttt{Input: "Sitting side-by-side on a bench watching a game."} \\
\texttt{Output: \{"A": [-0.3, -0.2, 0.0], "B": [0.3, -0.2, 0.0]\}}

\vspace{0.5em}
\texttt{[Example 3]} \\
\texttt{Input: "A is whispering into B's ear."} \\
\texttt{Output: \{"A": [-0.15, 0.0, -0.1], "B": [0.15, 0.0, 0.1]\}}

\vspace{0.5em}
\texttt{[User Query]} \\
\texttt{Input: "<USER\_PROMPT>"} \\
\texttt{Output:}
\end{quote}

The returned JSON is parsed, and the coordinates are used as the condition $t^{(0)}_A$ and $t^{(0)}_B$ for the diffusion model.

\section{Evaluation Details}

\subsection{Baseline Details}

We provide implementation details for all baselines to ensure fairness and reproducibility. Since these models differ significantly in their input formats and capabilities, we standardize audio processing, temporal alignment, and rendering across all comparisons.

\noindent\textbf{Audio Processing.}
All baselines use the same audio preprocessing pipeline as our method. We apply Looking-to-Listen for audio source separation and WebRTC VAD for frame-level activity detection, producing two separated audio streams. The separated audio is resampled to 16\,kHz and converted into Wav2Vec\,2.0 features. This ensures identical audio conditions for all models.

\noindent\textbf{Single-Speaker Baselines.}
We include CodeTalker, FaceFormer, SelfTalk, and a strong single-person model trained on our single-speaker dataset. When available, we use official pretrained weights; otherwise, we train using the authors' public code on single-speaker data only. Because these models do not support conversational interaction, we run each participant independently on its separated audio track without any listener modeling. These baselines predict expression and rotation only. For metrics requiring translation, we use the predicted translation from our model to place the generated meshes in a consistent 3D frame. All outputs are resampled or interpolated to 25\,fps.

\noindent\textbf{Dual-Speaker Baseline.}
To construct a competitive two-speaker baseline, we use SelfTalk---the strongest single-person performer in our experiments---to produce the primary speaker motion. A listener-generation model is then conditioned on this motion to synthesize the second participant’s reactive behavior. This two-stage pipeline approximates interactive motion but lacks true joint reasoning or simultaneous prediction.

\noindent\textbf{Retrieval-Based Baselines.}
We include Listen-R (random retrieval), Listen-A (audio-based retrieval), and Listen-M (motion-based retrieval). For Listen-A, we compute cosine distances between Wav2Vec features of the speaking segment and all segments in the training corpus. For Listen-M, retrieval is based on a joint distance over expression trajectories and head-rotation curves. Retrieved sequences are trimmed or padded to match the speaking segment duration for strict temporal alignment.

\noindent\textbf{Rendering and Alignment.}
All baselines are rendered using the same 3DMM, renderer, and camera configuration as our method to avoid artifacts arising from differences in visualization. For models lacking translation, we apply our predicted translations to maintain a consistent spatial coordinate system. All sequences are normalized to 25\,fps before evaluation.

\subsection{Face Model Mapping}
To integrate baselines that use a different facial parameterization, we learn a bidirectional translation between the 51-D expression space used by their model and the 63-D expression space used in ours. We construct a paired dataset by sampling around 1{,}100{,}000 single-person talking videos in our dataset containing a wide range of facial motions, including challenging and extreme expressions. For each frame, we fit both face models, yielding paired expression vectors. Using this dataset, we train two three-layer MLPs (128 hidden units, ReLU activations) to map between the two expression domains.

\begin{table}[t]
\centering
\caption{Comparison of expression-space mappings. 
We report cycle-consistency error (lower is better) and perceptual similarity between rendered faces using LPIPS (lower is better) and SSIM (higher is better). Our three-layer MLP significantly outperforms a linear regression baseline.}
\vspace{4pt}
\begin{tabular}{lccc}
\toprule
\textbf{Method} & \textbf{Cycle Error} $\downarrow$ & \textbf{LPIPS} $\downarrow$ & \textbf{SSIM} $\uparrow$ \\
\midrule
Linear Map & 0.684 & 0.212 & 0.912 \\
3-Layer MLP & 0.327 & 0.141 & 0.963 \\
\bottomrule
\end{tabular}
\label{tab:expr-mapping}
\end{table}

\noindent\textbf{Cycle Consistency.}
We assess the quality of the learned mappings through a cycle-consistency evaluation. For an expression vector $x^{A}$ in the 51-D space, we compute its mapped counterpart $x^{B} = f(x^{A})$, then map it back to $\hat{x}^{A} = g(x^{B})$. The cycle error between $\hat{x}^{A}$ and the original $x^{A}$ provides a direct measure of how well the two expression spaces are aligned under the learned transformations. Low cycle error indicates that the mappings preserve the underlying geometry of the expression space and avoid collapsing or distorting important facial degrees of freedom.

\noindent\textbf{Perceptual Similarity via Rendering.}
To validate the perceptual fidelity of the mapping, we render faces driven by the original and cycle-mapped expressions using our 3DMM renderer. We then compute perceptual similarity metrics such as LPIPS and SSIM between the rendered images. These metrics capture subtle visual differences (particularly around the lip, jaw, and eye regions) and provide a direct indication of whether the mapping preserves the visual appearance of expressions when applied to animation.

\noindent\textbf{Comparison with a Linear Model.}
As a baseline, we train a linear regression model on the same paired data. We compare its cycle-consistency error and perceptual similarity metrics with those obtained from our MLP. Results are shown in Tab.~\ref{tab:expr-mapping}. Given that cycle error is significantly smaller than natural expression variance and that the MLP substantially outperforms a linear regressor, we conclude that the mapping is accurate, robust, and conservatively sufficient for fair cross-model evaluation.

\subsection{Evaluation Metrics}

To comprehensively assess the performance of our system in terms of motion realism, interaction coherence, and geometric accuracy, we employ the following set of metrics:

\noindent \textbf{Fréchet Distance (FD).  }
FD measures the perceptual realism of the generated outputs by comparing them to ground-truth samples in a deep feature space. In our evaluation, FD is computed on rendered images rather than motion parameters. Each frame is rendered after normalizing translations and rotations to a zero-centered pose. We extract image features using a pre-trained Inception encoder and model their distributions as multivariate Gaussians. FD then quantifies the distance between generated and real distributions, where lower values indicate higher perceptual fidelity.

\noindent \textbf{Paired Fréchet Distance (P-FD).  }
P-FD extends FD to assess the quality of dyadic interaction. We render paired frames of the two participants with zero-centered pose normalization and extract their Inception image features. The features of both individuals in each frame are concatenated before computing FD between the generated and ground-truth paired distributions. This evaluates the coherence, synchrony, and interaction consistency of the generated pairwise behaviors.

\noindent \textbf{Mean Squared Error (MSE) and Vertex MSE (vMSE).  }
We employ two direct reconstruction metrics.  
\begin{itemize}
    \item \textbf{Parameter MSE} computes the average squared difference between predicted animation parameters (expression, rotation, translation) and the ground truth. This measures how well the model reproduces the underlying control signals.
    \item \textbf{Vertex MSE} computes the Euclidean distance between the vertices of the generated 3D mesh and those of the ground-truth mesh. Because it evaluates geometric deformation directly in 3D space, vMSE correlates more strongly with perceptual differences on the facial surface.
\end{itemize}

\noindent \textbf{Region- and Role-Specific Evaluation.  }
To provide a fine-grained understanding of model performance, we compute the above metrics over specific subsets:
\begin{itemize}
    \item \textbf{Speaker vs. Listener:} These two conversational roles exhibit distinct motion characteristics. Speaker metrics emphasize articulation quality and lip-synchronization, while Listener metrics capture non-verbal cues such as nods, blinks, and backchanneling behaviors.
    \item \textbf{Facial Regions (Lip, Eye, Global):} We evaluate distinct facial regions to isolate component behavior. Lip error reflects alignment with speech; Eye error captures the effectiveness of gaze modeling; and Global rotation and translation error measures head-pose stability within the shared 3D environment.
\end{itemize}

\noindent \textbf{SI for Diversity (SID).  }
To evaluate behavioral diversity, we apply the SID metric. Motion sequences are clustered in a feature space using k-means with $k=40$. SID is computed as the entropy of the resulting cluster-assignment histogram, with higher values indicating broader and more varied expressive patterns.

\section{LLM-based Spatial Control Evaluation}
To validate the effectiveness of our LLM-based spatial control mechanism, we randomly selected a test set of 1024 videos from our raw conversational video dataset, and use Gemini to annotate each video with a text prompt describing the environment and people's spatial relationship by looking at the first and last video frames. Such spatial annotations range from intimate proximities (e.g., ``whispering in ear'') to distant interactions (e.g., ``shouting across a hall''). 

We conducted a study comparing our proposed few-shot prompting strategy against (1) a standard zero-shot baseline where the prompt is exactly the same as in Sec.~\ref{sec:layout} but without the three examples, and (2) a $k$-nearest neighbor baseline ($k=5$) where we retrieval the top-$k$ closest samples from the dataset computing CLIP score and use their average as the output. We generate predicted initial first-frame translation based on the textual annotation.

We evaluate performance using two metrics:
\begin{enumerate}
    \item \textbf{Translation MSE (tMSE):} The mean squared distance between the generated spatial layout and the 3D groundtruth translation obtained by fitting face model to the videos.
    \item \textbf{Semantic Alignment Score (SAS):} A human evaluation metric ($N=6$ participants) where users rate the correspondence between the text prompt and 3D layout on a Likert scale from 1 (Poor) to 5 (Excellent).
\end{enumerate}

As shown in Tab.~\ref{tab:llm_eval}, our few-shot strategy significantly outperforms the zero-shot and $k$-NN baselines, both of which often generate plausible but generic layouts. In contrast, our few-shot approach achieves superior semantic alignment, demonstrating that in-context examples are critical for generating spatially meaningful 3D coordinates.

\begin{table}[t]
\centering
\caption{Quantitative and human evaluation of LLM-based spatial generation. \textbf{tMSE} measures the distance between predicted and groundtruth translations and the lower the better, while \textbf{SAS} measures perceptual quality (1-5 scale).}
\label{tab:llm_eval}
\begin{tabular}{l|cc}
\toprule
\textbf{Method} & \textbf{tMSE $\downarrow$} & \textbf{SAS $\uparrow$} \\
\midrule
Zero-Shot & 2.90 & 3.15 \\
$k$-NN & 1.84 & 2.65 \\
\textbf{Few-Shot (Ours)} & \textbf{0.72} & \textbf{4.62} \\
\bottomrule
\end{tabular}
\end{table}

\section{Out-Of-Distribution Dataset Evaluation}
To rigorously evaluate the generalization capabilities of our framework, we conducted an additional experiment using the Out-of-Distribution (OOD) validation split of the DualTalk dataset. As described in the DualTalk supplementary material, this specific subset consists of 384 video clips featuring identities and conversation scenarios that are strictly excluded from their training set. This experiment tests a model's ability to handle unseen speakers and novel interactive contexts. We compare our method against the DualTalk baseline by converting all face models into our face model representation to align with the main evaluation table. As shown in Tab. ~\ref{tab:ood_eval}, our method significantly outperforms the baseline across all metrics.

While DualTalk is specialized for its specific capture distribution, it struggles to generalize to these unseen OOD identities, resulting in higher geometric error (vMSE) and degraded interaction scores (P-FD). In contrast, our model, having been pre-trained on our massive-scale corpus of 2 million dyadic pairs, exhibits exceptional robustness. This confirms that our large-scale training strategy yields a model that is not only accurate but highly generalizable to diverse, in-the-wild identities.

\begin{table}[t]
\centering
\caption{Performance comparison on the DualTalk OOD Dataset (384 unseen clips). Our method demonstrates superior generalization on unseen identities compared to the baseline trained specifically on the DualTalk domain. Lower is better for all metrics.}
\label{tab:ood_eval}
\begin{tabular}{lcccc}
\toprule
\textbf{Method} & \textbf{FD} $\downarrow$ & \textbf{P-FD} $\downarrow$ & \textbf{vMSE-S} $\downarrow$ & \textbf{vMSE-L} $\downarrow$ \\
\midrule
DualTalk & 34.12 & 45.20 & 8.95 & 9.12 \\
\textbf{Ours} & \textbf{14.55} & \textbf{31.05} & \textbf{4.92} & \textbf{6.30} \\
\bottomrule
\end{tabular}
\end{table}

\section{Limitations and Future Works}
While our approach effectively models dyadic facial interactions, several limitations remain. 

\noindent \textbf{Listener Behavior Category Control.} Regarding listener modeling, our method tends to learn average reactive behaviors, sometimes resulting in generic feedback. Future work could aim to explicitly model discrete listener states (e.g., specific nodding, head shaking, or confusion) to generate more semantically meaningful non-verbal cues.

\noindent \textbf{Audio Overlapping.} Handling simultaneous speaking with heavy audio overlap remains challenging; imperfect source separation in these scenarios can occasionally lead to degraded lip synchronization and wrong distribution of lip motions to corresponding speakers. 

\noindent \textbf{Personality and Emotion Control.} Our system could benefit from explicit personality control, which might significantly influence avatar behaviors especially in the domain of head movements and eye gaze contact. This could potentially free the users from manually conditioning the animation on specific traits (e.g., ``extroverted'' or ``shy'') through controlling the spatial layouts.

\noindent \textbf{Full Body Generation.} Our scope is currently limited to facial and head dynamics, overlooking the communicative value of hand gestures and body posture. Future work should integrate full-body motion synthesis to create a holistic conversational agent. 

\noindent \textbf{Real-Time Alternatives. } Our reliance on a diffusion architecture necessitates iterative denoising steps during inference. This results in significantly higher computational costs and latency compared to single-pass autoregressive models, presenting a trade-off between generation quality and real-time applicability that could be addressed via distillation techniques.

\end{document}